\documentclass[10pt,journal]{IEEEtran}

\usepackage{cite}
\usepackage{amsmath,amssymb,amsfonts}
\usepackage{algorithmic}
\usepackage{graphicx}
\usepackage{textcomp}

\usepackage{graphicx}
\usepackage{float}
\usepackage{subfig}
\usepackage{subfloat}
\usepackage{booktabs}
\usepackage{multirow}
\usepackage{makecell}

\begin{document}

\title{Guidance Design for Escape Flight Vehicle Using Evolution Strategy Enhanced Deep Reinforcement Learning}

\author{Xiao Hu, Tianshu Wang, Min Gong, Shaoshi Yang,~\IEEEmembership{Senior Member,~IEEE}
\thanks{X. Hu is with the School of Aerospace Engineering, Tsinghua University, Beijing 100084, China, and also with China Academy of Launch Vehicle Technology, Beijing 100076, China (e-mail:  huxiao18@mails.tsinghua.edu.cn).}
\thanks{T. Wang is with the School of Aerospace Engineering, Tsinghua University, Beijing 100084, China (e-mail: tswang\_tsinghua@163.com).}
\thanks{M. Gong is with China Academy of Launch Vehicle Technology, Beijing 100076, China (e-mail: gongmin913@163.com).}
\thanks{S. Yang is with the School of Information and Communication Engineering, Beijing University of Posts and Telecommunications, with the Key Laboratory of Universal Wireless Communications, Ministry of Education, and also with the Key Laboratory of Mathematics and Information Networks, Ministry of Education, Beijing 100876, China (e-mail: shaoshi.yang@bupt.edu.cn).}
\thanks{\textit{Corresponding author: S. Yang}} 
}



\markboth{IEEE Access, accepted 18 March 2024, DOI: 10.1109/ACCESS.2024.3383322}%
{Shell \MakeLowercase{\textit{et al.}}: Bare Demo of IEEEtran.cls for IEEE Journals}

\maketitle

\begin{abstract}Guidance commands of flight vehicles can be regarded as a series of data sets having fixed time intervals, thus guidance design constitutes a typical sequential decision problem and satisfies the basic conditions for using the deep reinforcement learning (DRL) technique. In this paper, we consider the scenario where the escape flight vehicle (EFV) generates guidance commands based on the DRL technique and the pursuit flight vehicle (PFV) generates guidance commands based on the proportional navigation method. Evasion distance is described as the minimum distance between the EFV and the PFV during the escape-and-pursuit process. For the EFV, the objective of the guidance design entails progressively maximizing the residual velocity, which is described as the EFV's velocity when the evasion distance occurs, subject to the constraint imposed by the given evasion distance. Thus an irregular dynamic max-min problem of extremely large-scale is formulated. In this problem, the time instant when the optimal solution (i.e., the maximum residual velocity satisfying the evasion distance constraint) can be attained is uncertain and the optimum solution is dependent on all the intermediate guidance commands generated before. For solving this challenging problem, a two-step strategy is conceived. In the first step, we use the proximal policy optimization (PPO) algorithm to generate the guidance commands of the EFV. The results obtained by PPO in the global search space are coarse, despite the fact that the reward function, the neural network parameters and the learning rate are designed elaborately. Therefore, in the second step, we propose to invoke the evolution strategy (ES) based algorithm, which uses the result of PPO as the initial value, to further improve the quality of the solution by searching in the local space. Extensive simulation results demonstrate that the proposed guidance design method based on the PPO algorithm is capable of achieving a residual velocity of 67.24 m/s, higher than the residual velocities achieved by the benchmark soft actor-critic and deep deterministic policy gradient algorithms. Furthermore, the proposed ES-enhanced PPO algorithm outperforms the PPO algorithm by 2.7\%, achieving a residual velocity of 69.04 m/s.
\end{abstract}


\section{INTRODUCTION}
I{\scshape n} modern air combat environment involving escape flight vehicles (EFVs) and pursuit flight vehicles (PFVs), guidance design has attracted substantial research interests, since it constitutes an effective solution to the flight vehicle escape-and-pursuit problem. From the perspective of the pursuit side, in general the existing methods are devoted to optimizing the zero effort miss (ZEM) performance \cite{1, 2, 3, 4}. However, they cannot be used directly by the EFV, because of the essential distinction between the purposes of PFV and EFV.

As far as improving the escape capability of flight vehicle is concerned, traditional methods rely on mathematical analysis to generate a time series of guidance commands. More specifically, some researchers advocated to design the guidance laws by taking advantage of the differential game strategy \cite{5, 6}, which requires a large amount of computing resources. Hence, this method is inappropriate for EFV that has limited computing resources. Liu et al. \cite{7} proposed a coverage-based cooperative guidance law to intercept strongly maneuverable vehicles by using multiple weakly maneuverable vehicles. Sinha et al. \cite{8} proposed a method to reduce the time-to-go error and design a guidance law to intercept a moving target with the lateral acceleration constraints. Yu et al. \cite{9} proposed a method to derive the high-precision analytical solution of fight time for a group of coordinated flight vehicles, and designed a longitudinal guidance law and a lateral guidance law to achieve the goal of simultaneous arrival in the presence of multiple no-fly zones. Marchidan et al. \cite{10} proposed a method based on local parameterized guidance vector fields to address the collision avoidance problem. Xu et al. \cite{11} proposed a method on the basis of distributed perception and decision-making to solve the obstacle avoidance problem for unmanned aerial vehicles (UAVs). Mujumdar et al. \cite{12} proposed a method relying on both the nonlinear geometric guidance and differential geometric guidance to avoid the reactive collision at low altitudes of UAVs.

Owing to the strong feature representation ability of deep neural networks, deep learning (DL) \cite{13} is well known as a powerful machine learning (ML) technique that has solved many challenging problems. Meanwhile, reinforcement learning (RL) \cite{14}, an effective solution to sequential decision problems, is capable of training agents through the feedback information provided by the environment. As a combination of DL and RL, deep reinforcement learning (DRL) has achieved remarkable success \cite{15, 16, 17} and widely used in competitive games. AlphaGo \cite{18}, developed by DeepMind, defeated top human players in the game of Go in 2016, and its upgraded version AlphaGo Zero \cite{19} defeated AlphaGo through a self-learning method in 2017. AlphaStar \cite{20}, developed again by DeepMind, defeated professional players in the real-time strategy video game of StarCraft II by 5--0 in 2018. Wukong artificial intelligence (AI) \cite{21}, developed by Tencent, defeated top human players in the multiplayer online battle arena (MOBA) game of Honor of Kings in 2019.

On one hand, the guidance commands of the flight vehicle can be regarded as a series of data sets having fixed time intervals. Therefore, guidance design is a typical sequential decision problem and satisfies the basic conditions for using the DRL technique. On the other hand, the computing resources required by neural networks are less than those of the differential game strategy, thus the neural network based guidance design can be handled online by vehicles. Because of the above reasons, guidance design based on the DRL technique has become an emerging topic of great significance. Zhang et al. \cite{22} proposed a method based on DRL to learn the proportional coefficient, for which the traditional proportional guidance method provides the prior information, and realized an adaptive variable proportional coefficient guidance method. Liang et al. \cite{23} proposed a model-enhanced DRL method that can be divided into two steps. Firstly, a guidance design method based on model predictive control (MPC) was proposed. Secondly, a deep neural network was invoked to approximate the dynamics prediction model, which is valuable for satisfying the unknown disturbance possibly taking place during the flight. Rajagopalan et al. \cite{24} analyzed the feasibility and advantages of generating guidance commands directly by using neural networks. Gaudet et al. \cite{25} proposed a guidance design method based on DRL, and demonstrated its advantage compared with the traditional proportional guidance method. Meyer et al. \cite{26} proposed a collision avoidance guidance design method based on DRL for unmanned surface vehicles. Wu et al. \cite{27} proposed an obstacle avoidance method based on the Q-Learning algorithm for autonomous underwater vehicles. Li et al. \cite{28} proposed a guidance law based on neural networks to overcome the long-standing contradiction between the accuracy and real-time requirements of existing numerical predictive guidance methods. Although the above contributions have achieved valuable results, they cannot be used in the vehicle escape-and-pursuit problem directly, because of the different optimization objectives and different application scenarios. Cheng et al. \cite{29} proposed an intelligent predictor-corrector entry guidance approach for hypersonic vehicles by using deep neural networks, and demonstrated that their method is capable of achieving the trajectory correction with an update frequency of 20 Hz and providing high-precision, safe, and robust entry guidance for hypersonic vehicles. Peng et al. \cite{30} proposed a state-following-kernel-based RL method and successfully applied it to a vehicle-target interception system. It was demonstrated by numerical simulations that their proposed  guidance algorithm is capable of attaining an effective intercept to the vehicle-target engagement and the performance is superior to the state-of-the-art methods. Moreover, the implementation of the DRL technique in the guidance design of the UAVs not only enhances trajectory optimization but also extends significant benefits to various other domains, including but not limited to, the quality of communication \cite{31,32}.

Against the above backdrop, in this paper we aim to employ the DRL technique and the evolution strategy (ES) to generate real-time guidance commands for the EFV in the presence of a PFV that generates its guidance commands based on traditional methods. Our novel contributions are summarized as follows.

\begin{enumerate}
\item
Diverging from the conventional UAV, the EFV is subject to a distinct constraint, namely the guidance commands of the EFV are not allowed to vary dramatically at neighboring time instants. Additionally, given the EFV's embedded computing limitation, it cannot afford high computational complexity. To satisfy these constraints, we propose a highly customized interaction structure between the environment and the EFV's agent that invokes the DRL technique. The inputs of the agent are constituted by the current relative position between the EFV and the PFV, the current relative velocity between the EFV and the PFV, and the previous guidance commands of the EFV, while the outputs of the agent are set as the difference of the EFV's guidance commands between current and previous time instants. Our design provides sufficient information for the agent to generate reasonably valued change rates of guidance commands, so that the current guidance commands are guaranteed to be executable.
\item
The purpose of this paper is to generate optimal guidance commands that enable the EFV to effectively elude the PFV while maximizing the residual velocity. To address the problem, a novel reward function is designed for the proximal policy optimization (PPO) algorithm, which is a representative DRL technique suitable for solving the problem considered, by taking into account the change rate of the line-of-sight (LOS) direction of the PFV, and the prospective states (i.e. evasion distance and residual velocity) based on the current guidance commands of the EFV. Since this design introduces extra physics knowledge, the neural network invoked for continuously generating the guidance commands according to various real-time situations of the EFV and the PFV, can be trained in a more efficient manner.
\item
Since it is empirically not a good practice to use the convergence of the training curve as the criterion for determining the solution of the flight vehicle escape-and-pursuit problem, we propose to invoke an ES approach to perform further optimization based on the local optimum solution obtained by the PPO algorithm. Therefore, larger residual velocity satisfying the evasion distance constraint has been achieved by combining the advantages of the PPO algorithm and the ES approach. This is because the PPO algorithm is capable of quickly finding out a local optimum solution in a large space and the ES approach is capable of refining the solution in a small range with high precision.
\end{enumerate}

The rest of this paper is organized as follows. In Section \ref{section_system_model_and_problem_formulation}, we first present the system model, and then formulate the problem and analyze its characteristics. In Section \ref{section_the_proposed_evolution_strategy_enhanced_deep_reinforcement_learning_method}, the proposed ES-enhanced scheme relying on the PPO algorithm is derived, along with the detailed design process. In Section \ref{section_simulation_results_and_discusions}, the simulation results of guidance design by using the PPO algorithm and the ES-enhanced PPO algorithm are presented and discussed. Our conclusions are drawn in Section \ref{section_conclusions}.

\section{System Model and Problem Formulation}\label{section_system_model_and_problem_formulation}
\subsection{System Model}\label{subsection_system_model}
\begin{figure}[tbp]
\centerline{\includegraphics[width=0.95\linewidth]{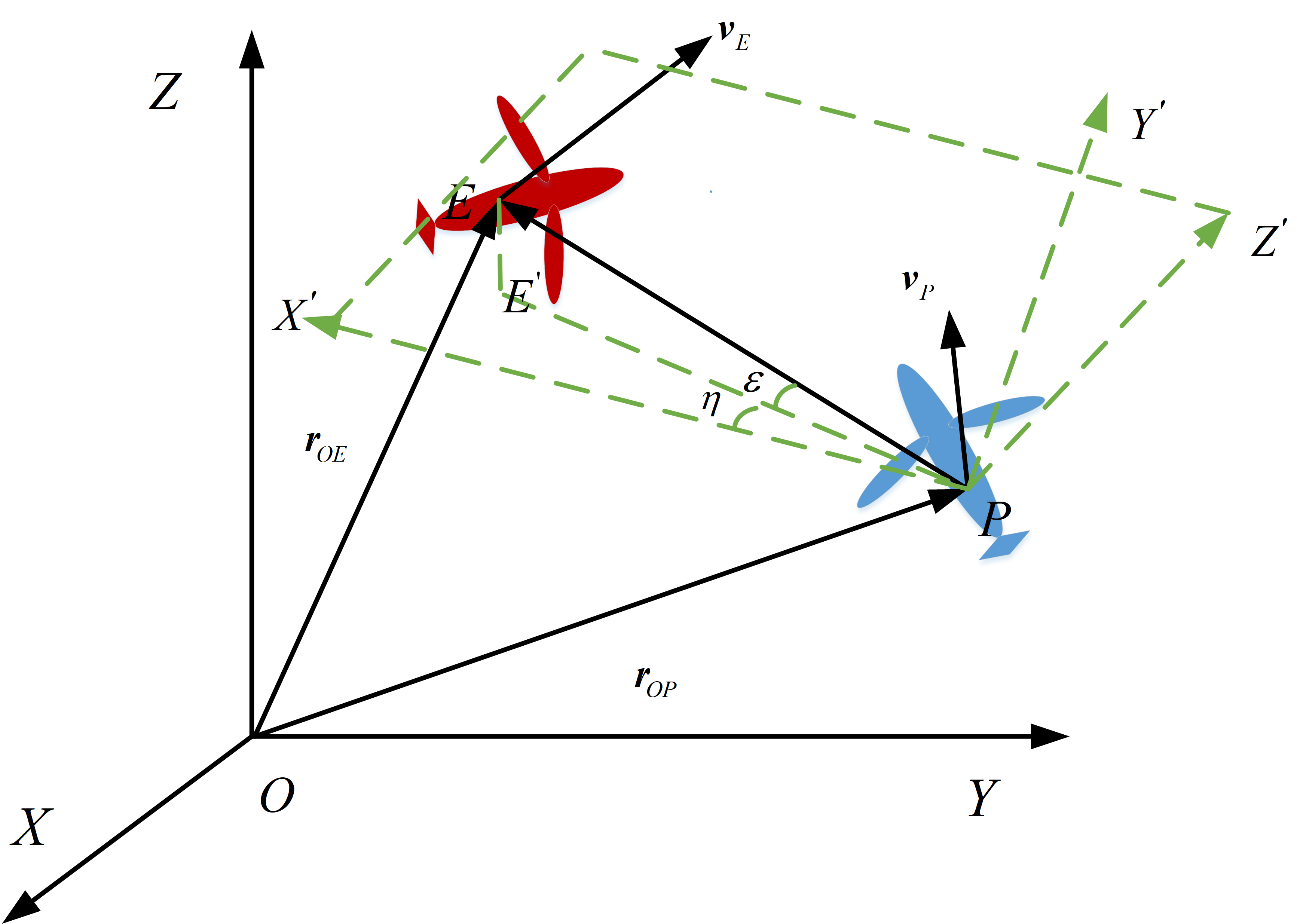}}
\caption{A combat scenario composed of a single EFV and a single PFV, where both the geocentric coordinate system ($X, Y, Z$) and a linearly shifted version of the launch coordinate system ($X', Y', Z'$) of the PFV are used.}
\label{fig1}
\end{figure}

We consider a combat scenario composed of a single EFV and a single PFV, where the EFV aims to generate its guidance commands with a judiciously designed algorithm, so that the optimization of evasion maneuver efficiency coincides harmoniously with the imperative of minimizing energy expenditure. Assume that the PFV generates its guidance commands based on the conventional proportional navigation method. For clarity, the combat scenario is illustrated with both the geocentric coordinate system ($X, Y, Z$) and a linearly shifted version of the launch coordinate system ($X', Y', Z'$) of the PFV, as shown in Figure \ref{fig1}.

Specifically, $O$ represents the center of the Earth and is also the origin of the geocentric coordinate system, while $OX$, $OY$ and $OZ$ are the three axes of the geocentric coordinate system. $E$ denotes the center of the mass of the EFV; $\boldsymbol{r}_{OE} = [{{r}_{x,E}}, {{r}_{y,E}}, {{r}_{z,E}}]$ is the position vector from the center of mass of the EFV to the Earth's center, and $\boldsymbol{v}_{E} = [{{v}_{x,E}}, {{v}_{y,E}}, {{v}_{z,E}}]$ is the velocity vector of the EFV. Additionally, $P$ denotes the center of the mass of the PFV; $\boldsymbol{r}_{OP} = [{{r}_{x,P}}, {{r}_{y,P}}, {{r}_{z,P}}]$ is the position vector from the center of mass of the PFV to the Earth's center, $\boldsymbol{v}_{P} = [{{v}_{x,P}}, {{v}_{y,P}}, {{v}_{z,P}}]$ is the velocity vector of the PFV. On the other hand, in the linearly shifted version of the launch coordinate system of the PFV, $E'$ is the projection of $E$ on the $X'-P-Z'$ plane, and $\varepsilon$ is the pitch angle from the LOS, and $\eta$ is the yaw angle from the LOS. The two angles can be described as
\begin{equation}
\begin{aligned}
 & {\varepsilon} = {f_1}({\boldsymbol{r}_{OP}}, {\boldsymbol{r}_{OE}}), \\
 & {\eta} = {f_2}({\boldsymbol{r}_{OP}}, {\boldsymbol{r}_{OE}}),
\end{aligned}
\label{eq1}
\end{equation}
where $f_1$ and $f_2$ are the abstract notation of certain functional relationships.

Furthermore, the following assumptions are made:
\begin{enumerate}
\item
Both the EFV and the PFV can accurately observe the present and historical position and velocity of each other, and can use the information to generate its own guidance commands. Nonetheless, the future position and velocity of them are hard to predict due to the interacting behavior of the EFV and the PFV.
\item
The EFV exhibits plane-symmetry, while its guidance commands are generated by an AI based method and constituted by the composite angle of attack\footnote{The composite angle of attack is given by ${\underline\alpha}_{\textrm{cx}} = \arccos(\cos{{\alpha }_{\textrm{cx}}}\cdot \cos{{\beta }_{\textrm{cx}}})$.} (denoted by ${\underline\alpha}_{\textrm{cx}}$) and the angle of heel (denoted by ${{\gamma}_{\textrm{cx}}}$). The range of ${\underline\alpha}_{\textrm{cx}}$ is $[-16.0^{\circ}, 16.0^{\circ}]$, and the range of ${\gamma}_{\textrm{cx}}$ is $[-90.0^{\circ}, 90.0^{\circ}]$.
\item
The PFV exhibits axial-symmetry, while its guidance commands are generated by the proportional navigation method \cite{33} and constituted by the angle of attack (denoted by ${\alpha}_{\textrm{cx}}$) and the angle of sideslip (denoted by ${{\beta}_{\textrm{cx}}}$). The range of both ${\alpha}_{\textrm{cx}}$ and ${\beta}_{\textrm{cx}}$ is $[-20.0^{\circ}, 20.0^{\circ}]$.
\item
The PFV has the capability to capture the EFV when the distance between them is less than 30.0m. More specifically, to evade capture by the PFV, the EFV must keep the distance larger than 30.0m during the escape-and-pursuit process.
\end{enumerate}

The proportional navigation method can be expressed as
\begin{equation}
\begin{aligned}
 & {{n}_{y'}} = {{K}_{1}} \frac{\textrm{d}\mathop{\varepsilon}}{\textrm{d}t}, \\
 & {{n}_{z'}} = {{K}_{1}} \frac{\textrm{d}\mathop{\eta}}{\textrm{d}t},
\end{aligned}
\label{eq2}
\end{equation}
where ${{n}_{y'}}$ is the command of the normal direction overload, ${{n}_{z'}}$ is the command of the lateral direction overload, $t$ denotes the time, and ${{K}_{1}}$ is the proportionality coefficient. Furthermore, we have
\begin{equation}
\begin{aligned}
 & {{\alpha}_{\textrm{cx}}} = {K}_{2} {{n}_{y'}},\\
 & {{\beta}_{\textrm{cx}}} = {K}_{2} {{n}_{z'}},
\end{aligned}
\label{eq3}
\end{equation}
where ${{K}_2}$ is the coefficient from the particular overload to its corresponding angle.

The reason why the proportional navigation method was chosen for the PFV is explained as follows. Common guidance methods for the PFV include: the proportional navigation method, guidance design based on differential games, and guidance design based on the DRL technique. If the PFV uses guidance design based on differential games, it would result in a significant computational load, which is unsuitable for PFV's application scenario. If the PFV uses guidance design based on the DRL technique, it would shift the system model of this paper from a single-agent model (only the EFV is an agent) to a multi-agent one (both the EFV and the PFV are agents), thus presenting a new problem. The purpose of this paper is to use an AI based method to generate the guidance commands of the EFV, so that the EFV's residual velocity satisfying evasion distance constraint can be maximized. Since the PFV can be regarded as a rival to facilitate the evaluation of the performance of the guidance command generating method designed for the EFV, it is sufficient to assume that the PFV uses the conventional proportional navigation method to generate its guidance commands. While some variations may arise if the PFV uses other guidance methods, it does not constitute the focus of this paper and can be investigated in our future work.

The vector form of the kinematics model of flight vehicles in the geocentric coordinate system is expressed as
\begin{equation}
\begin{aligned}
 & {{m}_{E}} \frac{{{\textrm{d}}^{2}}{\boldsymbol{r}_{OE}}}{\textrm{d}{{t}^{2}}} = {\boldsymbol{F}_{E}} + {\boldsymbol{R}_{E}} + {{m}_{E}} \boldsymbol{{g}_{E}},  \\
 & {{m}_{P}} \frac{{{\textrm{d}}^{2}}{\boldsymbol{r}_{OP}}}{\textrm{d}{{t}^{2}}} = {\boldsymbol{F}_{P}} + {\boldsymbol{R}_{P}} + {{m}_{P}} \boldsymbol{{g}_{P}},  \\
\end{aligned}
\label{eq4}
\end{equation}
where ${m}_{E}$ is the mass of the EFV, $\boldsymbol{F}_{E} = [{F}_{x,E}, {F}_{y,E}, {F}_{z,E}]$ is the EFV's control force vector with each element being a function of ${\underline\alpha}_{\textrm{cx}}$ and ${\gamma}_{\textrm{cx}}$, $\boldsymbol{R}_{E} = [{R}_{x,E}, {R}_{y,E}, {R}_{z,E}]$ is the EFV's aerodynamic force vector with each element also being a function of ${\underline\alpha}_{\textrm{cx}}$ and ${\gamma}_{\textrm{cx}}$, and $\boldsymbol{g}_{E} = [{{g}_{x,E}}, {{g}_{y,E}}, {{g}_{z,E}}]$ is the EFV's acceleration vector of gravity, whose elements are the functions of $\boldsymbol{r}_{OE}$. In addition, ${m}_{P}$ is the mass of the PFV, $\boldsymbol{F}_{P} = [{F}_{x,P}, {F}_{y,P}, {F}_{z,P}]$ is the PFV's control force vector, $\boldsymbol{R}_{P} = [{R}_{x,P}, {R}_{y,P}, {R}_{z,P}]$ is the PFV's aerodynamic force vector, and $\boldsymbol{g}_{P} = [{{g}_{x,P}}, {{g}_{y,P}}, {{g}_{z,P}}]$ is the acceleration vector of gravity. Each element of $\boldsymbol{F}_{P}$ and $\boldsymbol{R}_{P}$ is a function of ${\alpha}_{\textrm{cx}}$ and ${\beta}_{\textrm{cx}}$, while each element of $\boldsymbol{g}_{P}$ is a function of $\boldsymbol{r}_{OP}$.

\subsection{Evasion Distance and Residual Velocity}\label{subsection_evasion_distance}
The time-varying relative distance vector between the EFV and the PFV is $\boldsymbol{d}(t)$, which is constituted by ${{d}_{x}(t)}$, ${{d}_{y}(t)}$, and ${{d}_{z}(t)}$. The value of the relative distance ${d}(t)$ is given by
\begin{equation}
\begin{aligned}
 & {d}_{x}(t) = {r}_{x,E}(t) - {r}_{x,P}(t),  \\
 & {d}_{y}(t) = {r}_{y,E}(t) - {r}_{y,P}(t),  \\
 & {d}_{z}(t) = {r}_{z,E}(t) - {r}_{z,P}(t),  \\
 & d(t) = \sqrt{{{d}_{x}(t)}^{2} + {{d}_{y}(t)}^{2} + {{d}_{z}(t)}^{2}}.
\end{aligned}
\label{eq5}
\end{equation}

\begin{figure}[tbp]
\center
\centerline{\includegraphics[width=0.95\linewidth]{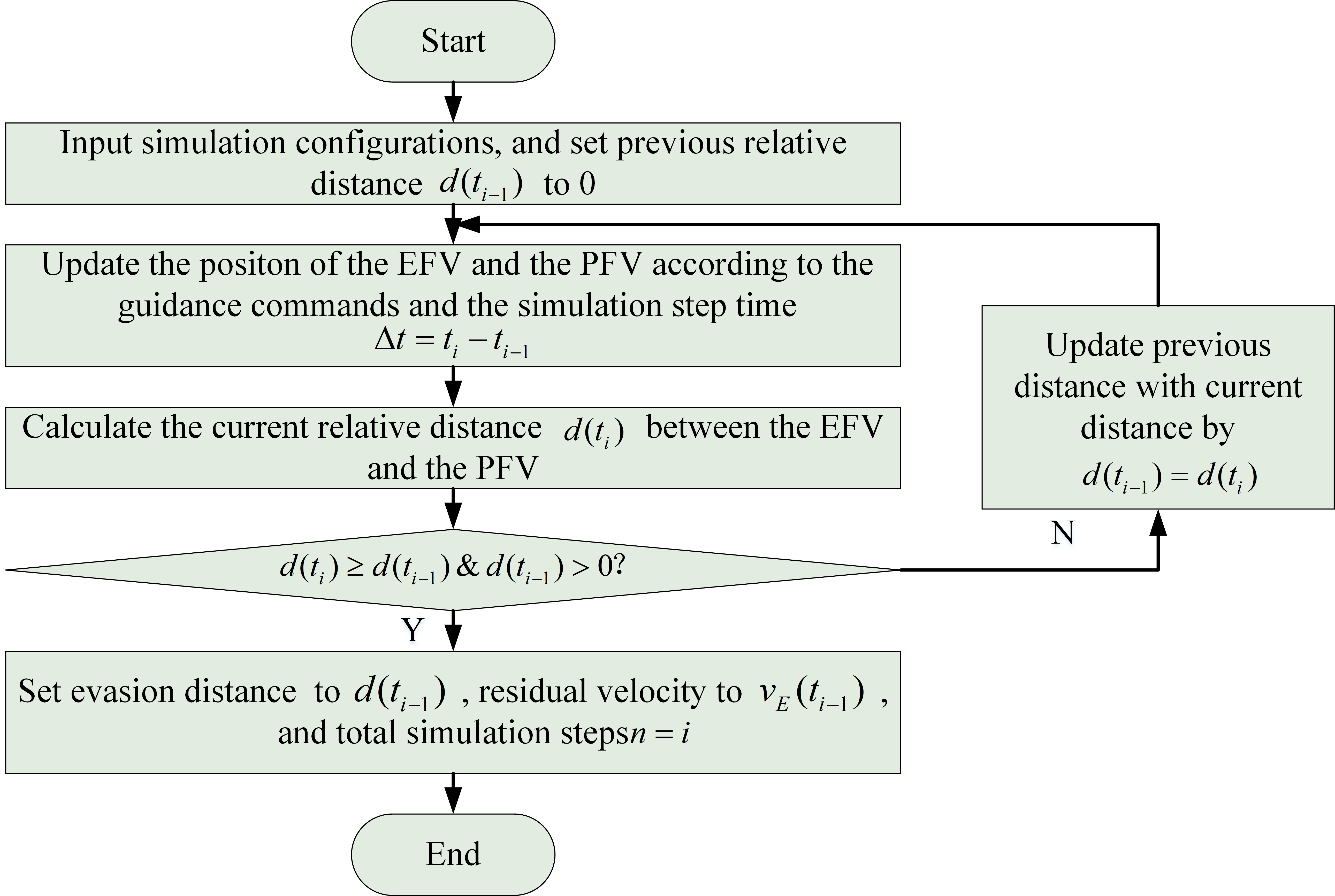}}
\caption{The calculation method of the evasion distance and the residual velocity.}
\label{fig2}
\end{figure}

The evasion distance is the minimum relative distance between the EFV and the PFV during the flight process, and the residual velocity is the velocity of the EFV when the evasion distance occurs (i.e., the time instant satisfies $d({{t}_{n}}) \ge d({{t}_{n-1}})$ and $d({{t}_{n-1}}) > 0$). The calculation method of the discrete-time evasion distance $d({t}_{i})$ and residual velocity ${v}_{E}({t}_{i})$ is shown in Figure \ref{fig2}, where $d({t}_{i})$ and ${v}_{E}({t}_{i})$ are the distance and EFV's velocity at the time instant $t_i$, respectively.

\subsection{Problem Formulation and Analysis}\label{subsection_problem_formulation_and_analysis}
Each single step of simulating the escape-and-pursuit scenario is described in Figure 3, and it consists of four major stages, i.e., Stage (1): The EFV generates its guidance commands based on its observation of the position and velocity of the PFV at previous time instant; Stage (2): According to the guidance commands generated by itself and its kinematics model, the EFV updates its own position and velocity at current time instant; Stage (3): The PFV generates its guidance commands based on its observation of the position and velocity of the EFV at previous time instant; Stage (4): Similarly, the PFV updates its position and velocity at current time instant according to the guidance commands generated by itself and its kinematics model. It should be noted that the current guidance commands of the EFV and PFV are based on observation of the position and velocity of each other at previous time instant, while the current position and velocity of the EFV and PFV are updated based on the current guidance commands of itself.

\begin{figure}[tbp]
\centerline{\includegraphics[width=0.95\linewidth]{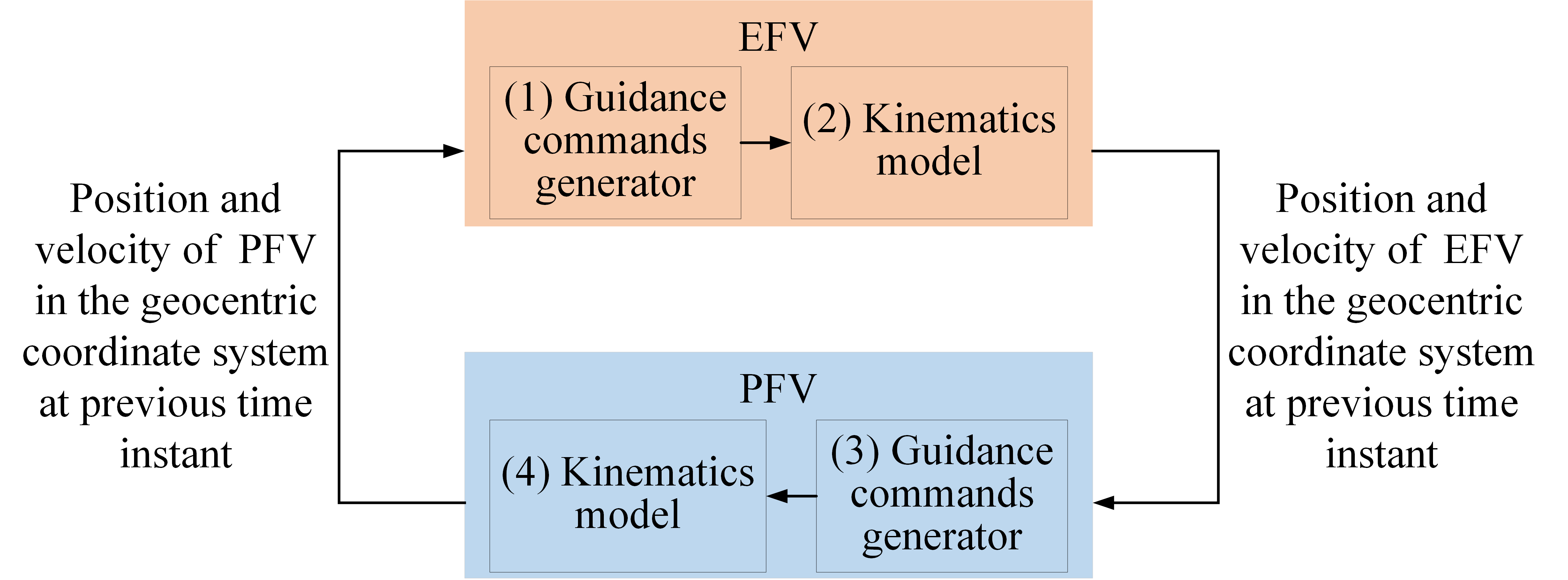}}
\caption{The iterative process of updating the positions and velocities of PFV and EFV.}
\label{fig3}
\end{figure}

Based on the above description, we can obtain the following insights:
\begin{enumerate}
\item
The kinematics model of the EFV is expressed by \eqref{eq4}, which means that the position and velocity of the EFV are readily available if the output of Stage (1) has been determined.
\item
According to its own position and velocity, as well as \eqref{eq1}, \eqref{eq2} and \eqref{eq3}, the guidance commands of the PFV are readily available if the position and velocity of the EFV have been determined.
\item
The kinematics model of the PFV is also expressed by \eqref{eq4}. Therefore, the position and velocity of the PFV are readily available if the output of Stage (3) has been determined.
\item
The only ``independent variable'' that can vary actively in each single step of simulating the escape-and-pursuit scenario, as illustrated by Figure \ref{fig3}, is the guidance command of the EFV, namely ${\underline\alpha}_{\textrm{cx}}$ and ${\gamma}_{\textrm{cx}}$, which constitute the output of Stage (1).
\end{enumerate}

As shown in Figure \ref{fig3}, the evasion distance can be obtained by using $d(t_{i-1}) = d(t_i)$ repetitively until the conditions $d(t_{i}) \ge d(t_{i-1})$ and $d(t_{i-1}) > 0$ are both satisfied. More specifically, $\boldsymbol d(t_i)$ is expressed as \eqref{eq6}.

\begin{figure*}
\begin{equation}
\begin{aligned}
 \boldsymbol d({{t}_{i}}) & = \boldsymbol d({{t}_{i-1}}) + f_d[({\underline\alpha}_{\textrm{cx}}({t}_{i}), {\gamma}_{\textrm{cx}}({t}_{i})] \\
 & = \boldsymbol d({{t}_{i-1}}) + [{\boldsymbol{v}_{E}({{t}_{i-1}})} - {\boldsymbol{v}_{P}({{t}_{i-1}})}]{\triangle}{t}
 + f_a[{\boldsymbol{r}_{OE}({{t}_{i-1}})}, {\boldsymbol{r}_{OP}({{t}_{i-1}})}, {\underline\alpha}_{\textrm{cx}}({t}_{i}), {\gamma}_{\textrm{cx}}({t}_{i}), {\alpha}_{\textrm{cx}}({t}_{i}), {\beta}_{\textrm{cx}}({t}_{i})]({\triangle}{t})^{2} \\
 & = \boldsymbol d({{t}_{i-1}}) + [{\boldsymbol{v}_{E}({{t}_{i-1}})} - {\boldsymbol{v}_{P}({{t}_{i-1}})}]{\triangle}{t} + \{[{{\boldsymbol{F}_{E}({\underline\alpha}_{\textrm{cx}}({t}_{i}), {\gamma}_{\textrm{cx}}({t}_{i}))} + {\boldsymbol{R}_{E}({\underline\alpha}_{\textrm{cx}}({t}_{i}), {\gamma}_{\textrm{cx}}({t}_{i}))} - {{m}_{E}}\boldsymbol{{g}}_{E}({\boldsymbol{r}_{OE}({{t}_{i-1}})}) }]/{{m}_{E}} \\
 & \quad - [{{\boldsymbol{F}_{P}({\alpha}_{\textrm{cx}}({t}_{i}), {\beta}_{\textrm{cx}}({t}_{i}))} + {\boldsymbol{R}_{P}({\alpha}_{\textrm{cx}}({t}_{i}), {\beta}_{\textrm{cx}}({t}_{i}))} - {{m}_{P}}\boldsymbol{{g}}_{P}( {\boldsymbol{r}_{OP}({{t}_{i-1}})})}]/{{m}_{P}}\} ({\triangle}{t})^{2},
\end{aligned}
\label{eq6}
\end{equation}
\end{figure*}

In \eqref{eq6}, ${\boldsymbol{v}_{E}({{t}_{i-1}})}$, ${\boldsymbol{v}_{P}({{t}_{i-1}})}$, ${\boldsymbol{r}_{OE}({{t}_{i-1}})}$ and ${\boldsymbol{r}_{OP}({{t}_{i-1}})}$ have all been determined by the guidance commands of the previous time instant $t_{i-1}$, ${\triangle}{t}$ is given by $t_i - t_{i-1}$, $f_d$ and $f_a$ are abstract notations of the functions corresponding to the distance and the acceleration, respectively. In addition, we have
\begin{equation}
\begin{aligned}
 & {\alpha}_{\textrm{cx}}({t}_{i}) = {{f}_{\alpha}[{\boldsymbol{r}_{OE}({{t}_{i-1}})}, {\boldsymbol{r}_{OP}({{t}_{i-1}})}]}, \\
 & {\beta}_{\textrm{cx}}({t}_{i}) = {{f}_{\beta}[{\boldsymbol{r}_{OE}({{t}_{i-1}})}, {\boldsymbol{r}_{OP}({{t}_{i-1}})}]}.
\end{aligned}
\label{eq7}
\end{equation}

\begin{figure}[tbp]
\centering
  \subfloat[The flight trajectories of the vehicles.]{
    \label{fig4a}
    \includegraphics[width=0.95\linewidth]{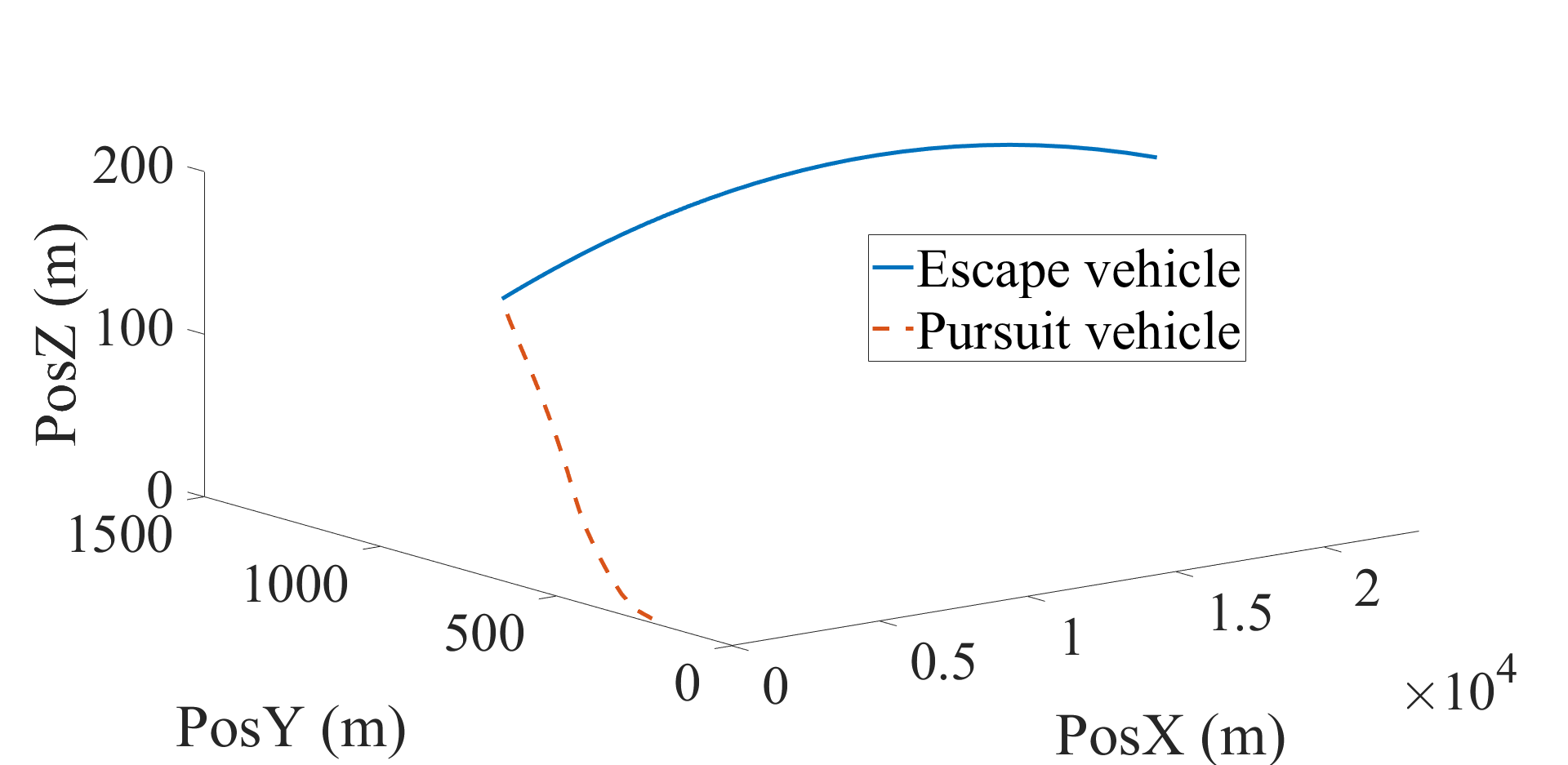}
  }
  \newline
  \subfloat[The relative distance of the vehicles.]{
    \label{fig4b}
    \includegraphics[width=0.95\linewidth]{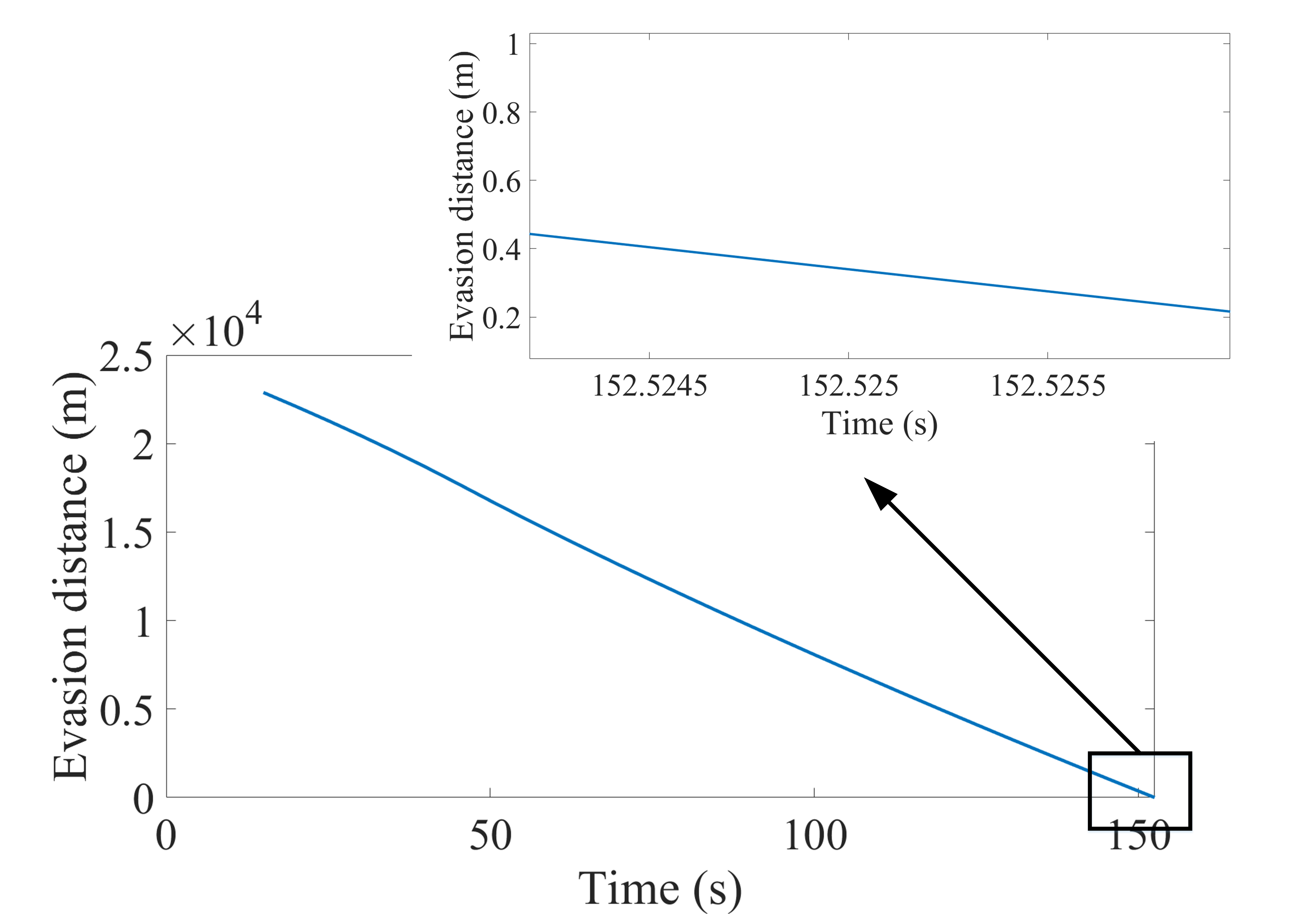}
  }
\caption{The simulation result without the maneuvering of the EFV.}
\label{fig4}
\end{figure}

In \eqref{eq7}, ${f}_{\alpha}$ and $f_{\beta}$ can be obtained by \eqref{eq1}, \eqref{eq2} and \eqref{eq3}. Following the above analysis, we conduct a simulation study, where the EFV ignores the PFV and takes no measure to avoid the PFV. The simulation result is shown in Figure \ref{fig4}. It can be observed from Figure \ref{fig4a} that the PFV flies to the EFV directly, and from Figure \ref{fig4b} that the evasion distance is as small as 0.2m, far shorter than 30.0m, which means the EFV is captured by the PFV completely in air combat.

Therefore, we conclude that it is necessary for the EFV to take advantage of its maneuverability proactively to avoid the PFV, and it is essential to study the guidance design method of the EFV, in order to obtain the maximum residual velocity satisfying the evasion distance constraint. Based on the previous discussions and derivations, this problem can be formulated as
\begin{equation}
\begin{aligned}
 \underset{{{\underline\alpha}_{\textrm{cx}}}({t}_{i}), {{\gamma}_{\textrm{cx}}}({t}_{i}), n} {\max} \quad  & {v}_{E}({{t}_{n-1}}), \\
 \textrm{s.t.}  \quad & \eqref{eq6}, \\
 & d({{t}_{n}}) \geq d({{t}_{n-1}}), \\
 & d({{t}_{n-1}}) > 30.0m, \\
 & {{\underline\alpha}_{\textrm{cx}}}({t}_{i}) \in [-16.0^{\circ}, 16.0^{\circ}], \\
 & {{\gamma}_{\textrm{cx}}}({t}_{i}) \in [-90.0^{\circ}, 90.0^{\circ}], \\
 & i \in [1, n].
\end{aligned}
\label{eq8}
\end{equation}

As seen from the problem \eqref{eq8}, the evasion distance is determined by the sequence of the guidance commands of the EFV. However, the specific time instant ${t}_{n}$ that satisfies $d({{t}_{n}}) \ge d({{t}_{n-1}})$ and $d({{t}_{n-1}}) > 0$ is uncertain, because it is influenced by the guidance commands generated at the previous time instants by the EFV. In addition, $n$ can be large under a sufficiently high computational precision. Moreover, the recursion relationship between $\boldsymbol d({{t}_{i}})$ and $\boldsymbol d({{t}_{i-1}})$, as characterized by \eqref{eq6}, is complicated. Last but not least, in aerodynamics, the explicit and accurate expressions of the various aerodynamic forces involved (i.e., $\boldsymbol{R}_{E}$ and $\boldsymbol{R}_{P}$) in terms of their respective independent variables, are usually unavailable. Therefore, it is difficult to find the analytical optimal solution of the problem \eqref{eq8}.

\section{The Proposed Evolution Strategy Enhanced Deep Reinforcement Learning Method}\label{section_the_proposed_evolution_strategy_enhanced_deep_reinforcement_learning_method}
In the traditional flight vehicle guidance designs, the input information is the state of the target, and the purpose of the generated guidance commands is taking the flight vehicle to the target continuously based on the state of the flight vehicle itself. For the proposed evasion guidance design, the input information of the EFV consists of the position and velocity of the PFV, but the objective is to maximize the EFV's residual velocity while maintaining a safe evasion distance. Although adopting a LOS-based guidance method is suitable for maximizing the evasion distance, it is not an optimal choice for maximizing the residual velocity of the EFV while keeping the safe evasion distance. Therefore, it is difficult to adapt the LOS-based guidance design method to the flight vehicle escape-and-pursuit problem considered.

Guidance commands of flight vehicles can be regarded as a series of data sets having fixed time intervals, thus guidance design constitutes a typical sequential decision problem and satisfies the basic conditions for using the DRL technique. The general idea of using DRL techniques to solve specific problems is to use the convergence of the training result based on the reward function as the criterion of the final solution. The drawback of the above approach arises when dealing with complex problems featuring a non-convex solution space, and the DRL technique may converge to the local optimum solution with the converged training result, and cannot obtain better solutions due to the absence of the ground-truth.

In order to reduce the risk of being trapped in a local optimum without realizing it, in this paper we propose a DRL based optimization strategy that is mainly composed of two steps as detailed below.
\begin{enumerate}
\item
In the first step, choose a suitable DRL technique to solve the escape-and-pursuit problem, based on judiciously designing the reward function, the neural network structure, and the learning rate, with the aid of domain knowledge.
\item
In the second step, invoke the output result of the DRL technique as the initial value, an evolution strategy based algorithm is carried out to further improve the result.
\end{enumerate}

\subsection{PPO Algorithm with Domain Knowledge for the escape-and-pursuit Problem}\label{subsection_ppo_algorithm_with_domain_knowledge_for_the_escape_and_pursuit_problem}
As discussed in Subsection \ref{subsection_problem_formulation_and_analysis}, the only independent variable in each single step of simulating the escape-and-pursuit scenario is the output of the Stage (1) in Figure \ref{fig3}, namely the guidance commands of the EFV. Because these guidance commands take their legitimate values from multi-dimensional continuous spaces, in principle the family of policy gradient algorithms can be the appropriate candidates of the solving method. Policy gradient algorithms are generally divided into two categories, namely the on-policy and the off-policy algorithms. The on-policy algorithms use a policy neural network to interact with the environment, so that the training data can be generated, which is then utilized to update the policy neural network itself immediately. Therefore, in on-policy algorithms, the obtained training data can only be used once. As a result, typically on-policy algorithms require a longer training time than off-policy algorithms. On the other hand, during the initial period of the training process, the policy neural network of both the on-policy and off-policy algorithms may be updated dramatically, because significant difference can exist between the training data obtained at neighboring episodes\footnote{The significant difference is due to exploration in a huge action space with a policy neural network yet to be optimized.}. In this case, it becomes difficult for the policy neural network to quickly find a good solution. To address this issue, the researchers from OpenAI proposed the PPO algorithm \cite{34}, which imposes constraints on the magnitude of the update carried out by the policy neural network. The PPO algorithm has been demonstrated effective in solving problems that are featured with multi-dimensional continuous action space. In what follows, we will employ the PPO algorithm to solve the problem considered.

\begin{figure}[tbp]
\centerline{\includegraphics[width=0.95\linewidth]{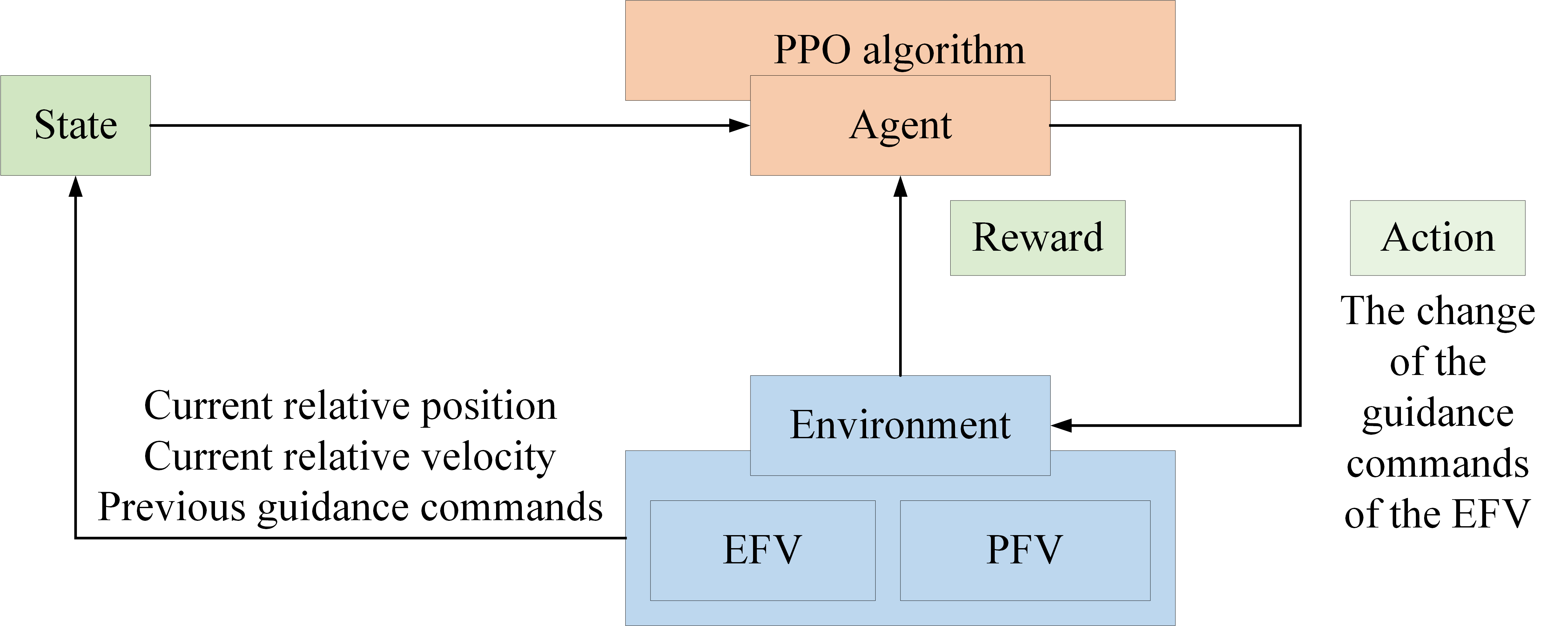}}
\caption{The interaction structure between the environment and the agent that invokes the PPO algorithm.}
\label{fig5}
\end{figure}

Then, the crucial work is to design the interaction structure between the environment and the agent, which can generate the guidance commands of the EFV based on the information obtained from the environment. It is worth noting that the explicit inputs of the Stage (1) in Figure \ref{fig3} are the position and velocity of the PFV, while the outputs of the Stage (1) also rely on the position and velocity of the EFV itself implicitly. As shown in Subsection \ref{subsection_system_model}, both the position and velocity of the EFV and PFV are described by six variables, respectively, namely ${{r}_{x,E}}$, ${{r}_{y,E}}$, ${{r}_{z,E}}$, ${{v}_{x,E}}$, ${{v}_{y,E}}$, ${{v}_{z,E}}$, and ${{r}_{x,P}}$, ${{r}_{y,P}}$, ${{r}_{z,P}}$, ${{v}_{x,P}}$, ${{v}_{y,P}}$, ${{v}_{z,P}}$. An intuitive idea is to set the above twelve variables as the inputs of the agent directly. However, the absolute values of the position and velocity of the EFV and PFV are not really meaningful for the escape-and-pursuit problem considered. Using these absolute values may cause the agent to treat the absolute values as the feature of the problem mistakenly, thus the generalization capability of the agent trained may be degraded. Therefore, it is a better alternative to set the relative position and relative velocity, totally six variables, as the inputs of the agent. As a result, the computational complexity can be reduced while improving the adaptability of the agent. In addition, owing to the limitations imposed by the physical property of the EFV, the guidance commands generated are not allowed to vary dramatically between neighboring time instants. In other words, the guidance commands at current time instant should only change in a reasonable range based on the guidance commands at previous time instant, rather than the entire range assumed in Subsection \ref{subsection_system_model}. Based on the above discussions, the inputs of the agent are composed of eight variables, of which six are the relative position and velocity between the EFV and PFV, and another two are the guidance commands ${\underline\alpha}_{\textrm{cx}}$ and ${{\gamma }_{\textrm{cx}}}$ at previous time instant. Correspondingly, the outputs of the agent are composed of two variables, namely the change of the guidance commands between current and previous time instants, as shown in Figure \ref{fig5}, where the interaction structure between the environment and the agent that invokes the PPO algorithm is illustrated.

In addition to the chosen PPO algorithm and the designed interaction structure, we need to design the reward function, the neural network and the learning rate, all of which have significant impact on the trained agent.

\subsubsection{Design of the Reward Function}\label{subsubsection_design_of_the_reward_function}
In the DRL technique, the reward value obtained from the environment determines the direction of the optimization of the agent, and the training purpose is to make the agent steadily obtain the highest possible reward. Therefore, the reward function that can accurately characterize the effect of the action currently taken subject to the current state is very important. We propose a domain-knowledge aided reward function expressed as
\begin{equation}
\begin{aligned}
 R = \sum{{R}_{p}} + {R}_{f},
\end{aligned}
\label{eq9}
\end{equation}
where $R$ is the total reward of a single training episode, ${R}_{p}$ is the immediate reward of every single step, and ${R}_{f}$ is the reward of the final step, when the evasion distance occurs, in the training episode. Since the goal of the optimization is to maximize the residual velocity satisfying the evasion distance constraint, the final step reward ${R}_{f}$ is expressed as
\begin{equation}
 {R}_{f} = {{K}_{v}} \times {{v}_{E}({t}_{n-1})} + {{K}_{d}} \times 30.0,
\label{eq10}
\end{equation}
where ${{K}_{v}}$ equals 10.0 when the evasion distance $d({t}_{n-1})$ exceeds 30.0m (i.e., the assumed safe evasion distance), otherwise it is set to 0.0, and ${{v}_{E}({t}_{n-1})}$ represents the residual velocity of the EFV at the time instant of ${t}_{n-1}$ when the conditions (i.e., $d({{t}_{n}}) \ge d({{t}_{n-1}})$ and $d({{t}_{n-1}}) > 0$) are satisfied in Figure \ref{fig2}. In addition, we have
\begin{equation}
{K}_{d} = \mathsf{clip}(d({t}_{n-1}) / 30.0, 0, 1),
\label{eq11}
\end{equation}
where $\mathsf{clip}(x, \min, \max)$ outputs $x$ if $\min \le x \le \max$, outputs $\min$ if $x < \min$, and outputs $\max$ if $x> \max$. Hence ${K}_{d}$ is limited to the range of $[0, 1]$. The presence of a discontinuous final reward introduces a significant level of complexity in solving the problem.

First of all, the number of simulation steps (i.e., the value of $n$ that satisfies $d(t_n) \ge d(t_{n-1})$ and $d(t_{n-1}) > 0$, as shown in Figure \ref{fig2} in a single episode is large. More specifically, the step time (i.e., $\Delta t$) used in the simulation must be set small, because the evasion distance obtained by solving the escape-and-pursuit problem, where the EFV and the PFV have a high relative speed, is sensitive to the step size. In the long-term learning problem (i.e., the problem that has a large $n$), it is very hard to learn from the sparse discontinuous reward (i.e., the total reward $R$ becomes $R_f$ when setting ${R}_{p}$ to zero; then the agent is trained with ${R}_{f}$ alone). Furthermore, since the step number $n$ is large and the immediate reward ${R}_{p}$ of every single step is added to the total reward $R$, any small error in $R_p$ may be accumulated large, leading to a failed training. Therefore, it is essential to design ${R}_{p}$ skillfully.

The immediate reward, while taking the current state of PFV and the future state of EFV into consideration, is expressed as
\begin{equation}
\begin{aligned}
 {R}_{p} = {R}_{p_e} + {R}_{p_v},
\end{aligned}
\label{eq12}
\end{equation}
where ${R}_{p_e}$ is the reward associated with the energy consumption of the PFV at the current time instant and ${R}_{p_v}$ is the reward associated with the prospective residual velocity of the EFV.

${R}_{p_e}$ is calculated by
\begin{equation}
 {R}_{p_e} = \frac{\textrm{d}\sqrt{{{\varepsilon}^{2}}+{{\eta}^{2}}}}{\textrm{d}t},
\label{eq13}
\end{equation}
where $\varepsilon$ is the pitch angle from the LOS, and $\eta$ is the yaw angle from the LOS, as introduced in Subsection \ref{subsection_system_model}. The main reason why $R_{p_e}$ is constructed as the expression of \eqref{eq13} is explained as follows. Firstly, an advantageous pursuit strategy of the PFV is the face-to-face hit \cite{35}, which means the motion direction of the PFV and the LOS direction of the PFV coincide. To this end, it is beneficial for the PFV to keep $\varepsilon$ and $\eta$ fixed in the escape-and-pursuit scenario. Secondly, because the change rate of $\varepsilon$ and $\eta$, as characterized by \eqref{eq13}, is positively related to the energy consumed by the PFV and the energy supply for the PFV is limited, it is beneficial for the EFV to make ${R}_{p_e}$ as big as possible.

\begin{figure}[tbp]
\centerline{\includegraphics[width=0.95\linewidth]{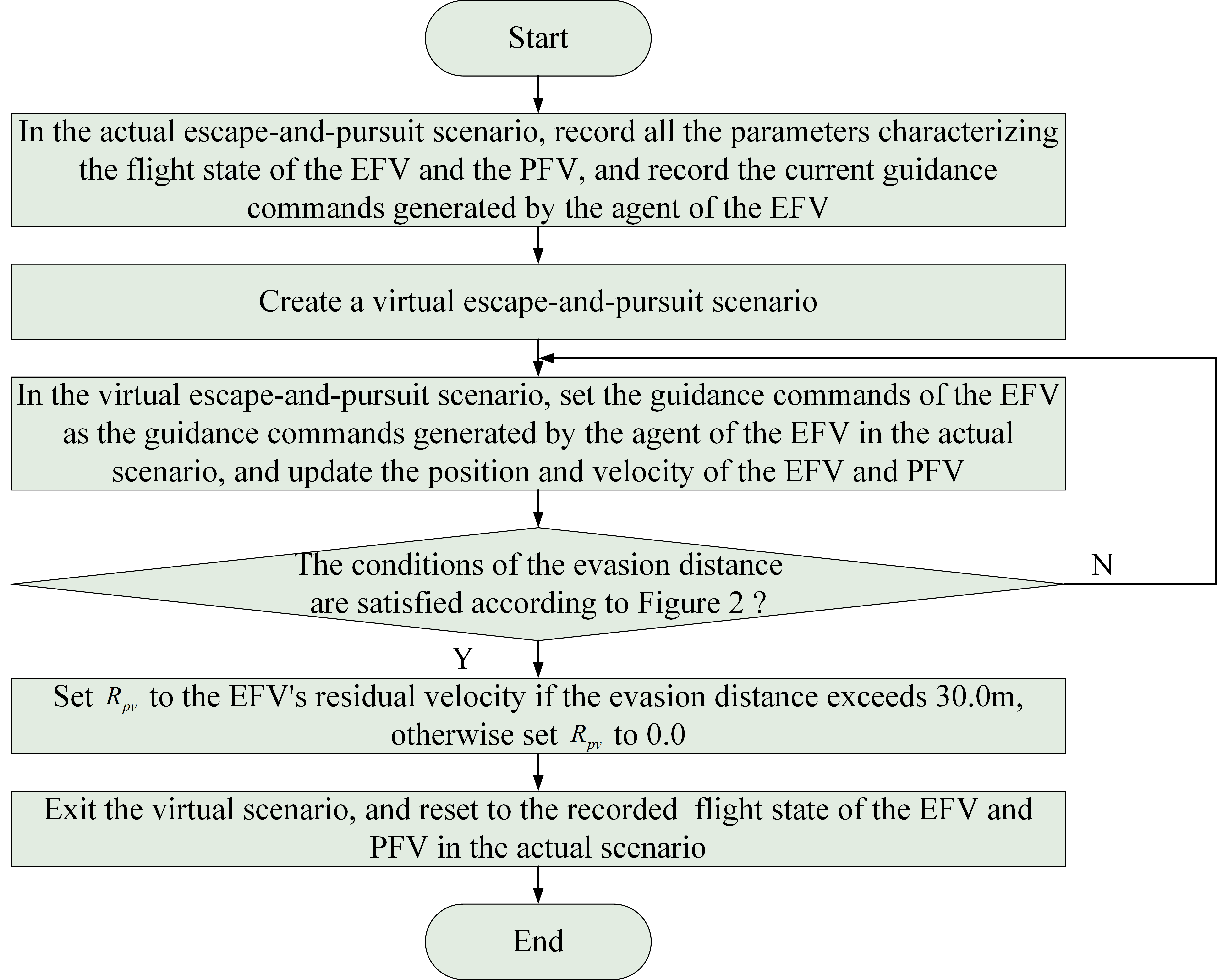}}
\caption{The procedure of calculating the prospective residual velocity and the prospective evasion distance in a virtual escape-and-pursuit scenario.}
\label{fig6}
\end{figure}

${R}_{p_v}$ is the function of the prospective residual velocity and the prospective evasion distance, which can be obtained through keeping the current guidance commands generated by the agent of the EFV fixed, until the end of the \textit{virtual escape-and-pursuit scenario} created. For clarity, this process is described in Figure \ref{fig6}.

\subsubsection{Design of the Neural Network}\label{subsubsection_design_of_the_neural_network}
\begin{figure}[tbp]
\centerline{\includegraphics[width=0.85\linewidth]{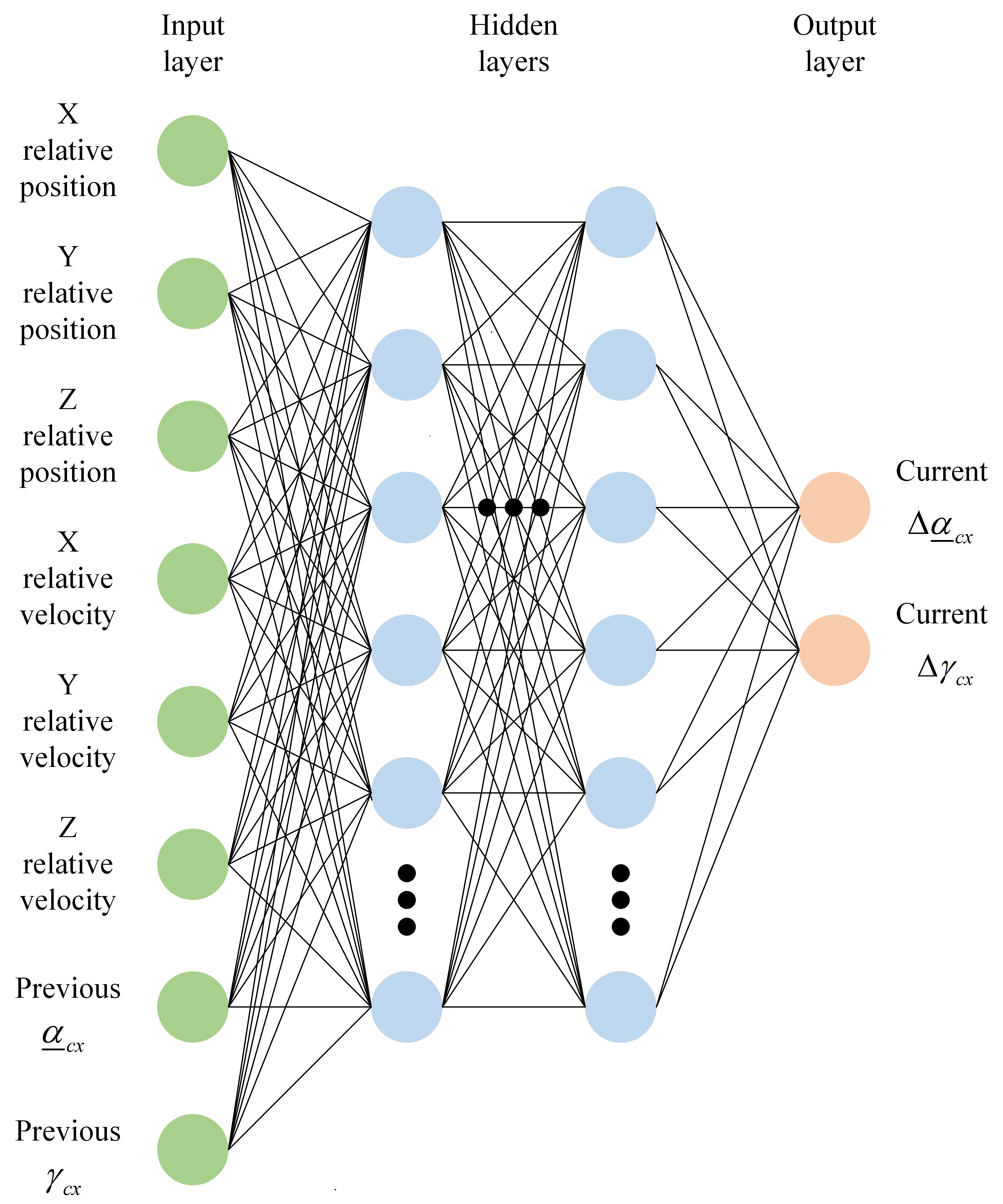}}
\caption{The architecture of the neural network specified by the hyper-parameter sets of Table \ref{tab1}. }
\label{fig7}
\end{figure}

With the same inputs, the outputs of an agent that is embedded in the EFV and relies on a neural network (also known as model) are determined mainly by three parts, namely the pre-processing of the neural network, the hyper-parameters of the neural network, and the post-processing of the neural network. Since the interaction structure between the environment and the agent has been designed and the PPO algorithm is chosen, as shown in Figure \ref{fig5}, both the pre-processing and the post-processing parts have been determined. Hence, it is of great importance to tune the neural network's hyper-parameters, which include but not limited to the number of layers, the number of nodes in each layer, and the specific activation functions\footnote{A fully connected architecture is assumed for the neural network to be designed, because all the inputs of the neural network are essential to its outputs in the particular problem considered.}. The degree of freedom for designing the neural network is large theoretically, but it is also computationally intensive. So far, no set rules exist to determine the optimal values of the hyper-parameters. Therefore, it is a pragmatic manner to tune the hyper-parameters from the empirically given multiple sets of hyper-parameters. The candidates of the hyper-parameter sets to be used in the experiments are shown in Table \ref{tab1}, and the illustration of the architecture corresponding to the neural network specified by Table \ref{tab1} is shown in Figure \ref{fig7}.

\begin{table}[tbp]
\centering
\caption{Candidates of the neural network's hyper-parameter set}
\begin{tabular}{cc}
\toprule
\textbf{Hyper-parameter}         &              \textbf{Value}         \\
\midrule
 \multirow{3}*{Architecture}     &  [8, 64, 64, 2], [8, 128, 128, 2], \\ & [8, 256, 256, 2], [8, 64, 64, 64, 2], \\ & [8, 128, 128, 128, 2], [8, 256, 256, 256, 2].  \\
 Activation function             &                 ReLU                \\
\bottomrule
\end{tabular}
\label{tab1}
\end{table}

To elaborate a little further, the number of nodes on the input layer and on the output layer is set to 8 and 2, respectively. The former equals the number of input variables of the agent, while the latter equals the number of output variables of the agent, as explicitly pointed out in the previous introduction of Figure \ref{fig5}. Let us consider [8, 256, 256, 256, 2] as an example, the neural network is composed of five layers, where the number of nodes on each hidden layer is set to 256. Furthermore, the total number of weight parameters describing the architecture of the neural network is 133632, which is obtained by $8 \times 256 + 256 \times 256 + 256 \times 256 + 256 \times 2$. As for the activation function, the reason for choosing ReLU is to reduce the computational complexity and make it suitable for the embedded system of the EFV.

\subsubsection{Design of the Learning Rate}\label{subsubsection_design_of_the_learning_rate}
The learning rate is another important hyper-parameter for tuning the neural network. It controls the magnitude of parameter updates by using the gradient based method during each iteration in response to the specific objective function. In DRL, since the reward is usually determined by the physical environment or the application, it may not be formulated as a differentiable function of the neural network's parameters. Hence, the gradient cannot be directly computed from the reward. In this paper we employ the PPO algorithm to establish the relationship between the reward and the objective function, which is differentiable and depends on the parameters characterizing the neural network. As a result, optimizing the objective function also maximizes the reward obtained by the agent.

The learning rate affects both the performance of the neural network and the convergence speed of the neural network's hyper-parameters. When training neural networks, it is often beneficial to reduce the learning rate as the training progresses. In this paper, a learning rate schedule based on a pre-defined function is adopted to provide the suitable learning rates, which are updated according to:
\begin{equation}
\begin{aligned}
 & {{P}_{R}} = \frac{{{S}_{T}}-{{S}_{E}}}{{{S}_{T}}}, \\
 & {{l}_{C}} = {{l}_{B}}  {{{{P}_{R}}}^{k}},   \\
\end{aligned}
\label{eq14}
\end{equation}
where ${S}_{T}$ is the specified total number of training steps, ${S}_{E}$ is the number of training steps having been executed, ${P}_{R}$ indicates the residual training progress, ${l}_{B}$ is the starting value of the learning rate, ${k}$ is the exponent, and ${l}_{C}$ is the actual learning rate invoked to update the neural network's hyper-parameters. The actual learning rate, which is calculated by \eqref{eq14}, is positively proportional to the residual training progress indicator $P_R$. Since the performance of the trained neural network embedded in the agent is expected to be positively proportional to the training process having been executed, the learning rate schedule adopted in this paper can strike a balance between improving the convergence speed and reducing the risk of getting stuck in local optimum. The parameter setup for updating the learning rate is shown in Table \ref{tab2}.

\begin{table}[tbp]
\centering
\caption{Parameter setup for updating the learning rate}
\begin{tabular}{cc}
\toprule
 \textbf{Parameter}                 & \textbf{Value}             \\
\midrule
 $k$                                &  2, 3                      \\
 ${l}_{B}$                          &  0.001, 0.0001             \\
 ${S}_{T}$                          &  3500000                   \\
\bottomrule
\end{tabular}
\label{tab2}
\end{table}

\subsection{ES-Enhanced PPO Algorithm}\label{subsection_es_enhanced_ppo_algorithm}
As pointed out in Subsection \ref{subsection_system_model}, compared with a conventional problem that has a multi-dimensional continuous-valued action space, the problem \eqref{eq8} considered has three additional features. The first one is the uncertain timing of achieving the evasion distance, the second one is that the number of simulation steps, $n$, may be large when optimizing the attainable evasion distance with fine-grained granularity, and the third one is that the reward is discontinuous as articulated in Subsubsection  \ref{subsubsection_design_of_the_reward_function}.

\begin{figure*}[tbp]
\centering
\includegraphics[width=0.8\linewidth]{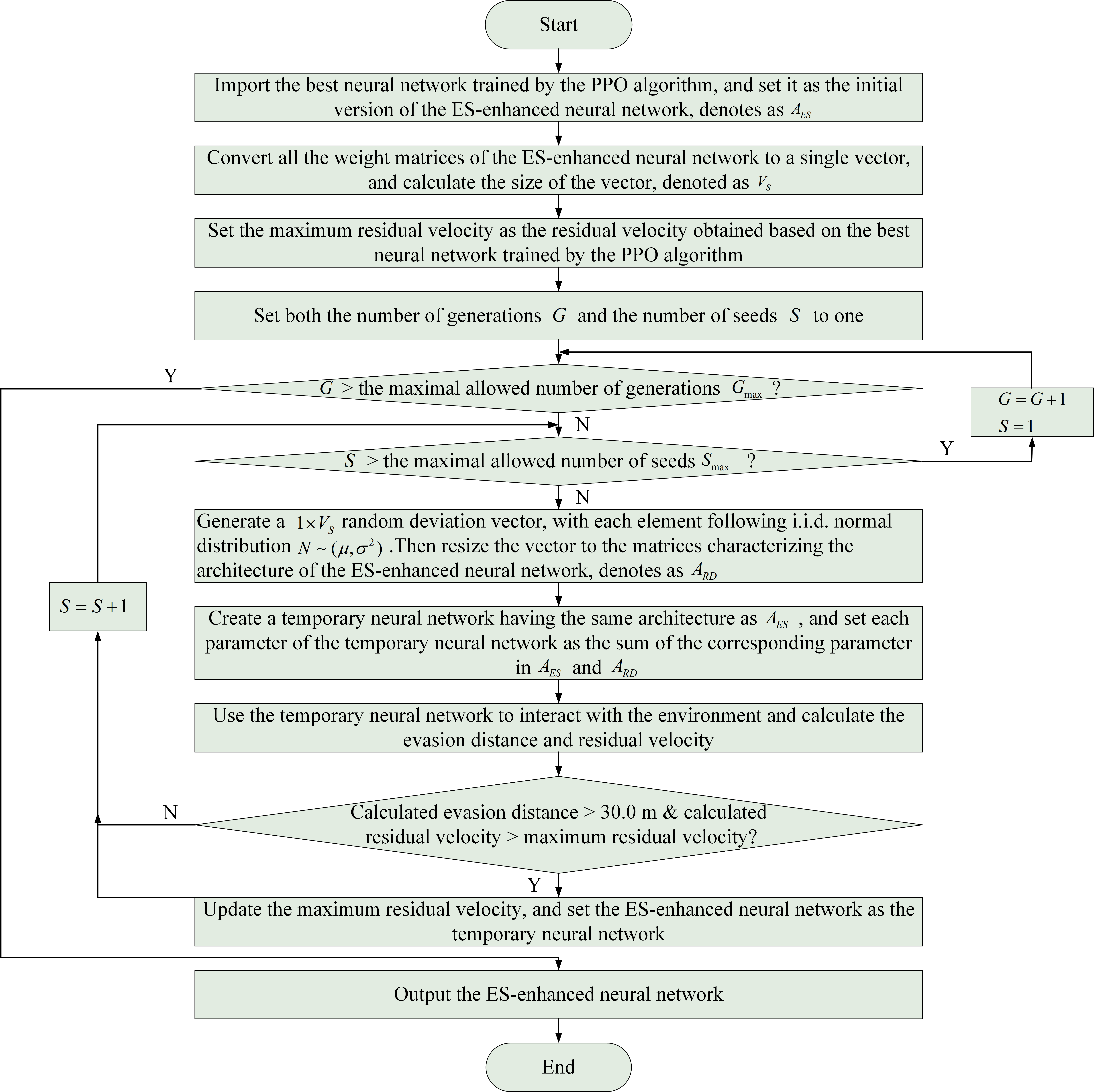}
\caption{The ES-enhanced PPO algorithm procedure.}
\label{fig8}
\end{figure*}

As the PPO algorithm updates the agent's neural network on the basis of the learning rate and the reward, it is hard to guarantee that the solution obtained for updating the weight parameters at the next time instant is always better than that obtained at the current time instant. It is well known that ES is effective in refining the solution obtained by other methods. Therefore, an ES-enhanced PPO algorithm is proposed to optimize the hyper-parameters of the neural network, and its specific procedure is shown in Figure \ref{fig8}, where the main parameters and their corresponding values used for optimizing the neural network's weight parameters are given in Table \ref{tab3}.

\begin{table}[tbp]
\centering
\caption{Parameters configurations of the ES-enhanced PPO algorithm}
\begin{tabular}{ccc}
\toprule
\textbf{Parameter}           &  \textbf{Meaning}                                                            &  \textbf{Value}                        \\
\midrule
$ G_{max} $                  &  \makecell{The maximal allowed \\ number of generations}                     &    50, 100, 150, 200                   \\
$ S_{max} $                  &  \makecell{The maximal allowed number \\ of seeds in each generation}        &    50, 100, 150, 200                   \\
$ \mu $                      &  The mean of normal distribution                                             &    0.0                                 \\
$ \sigma $                   &  The variance of normal distribution                                         &    0.01, 0.1, 1.0                      \\
\bottomrule
\end{tabular}
\label{tab3}
\end{table}

\section{Simulation results and discussions}\label{section_simulation_results_and_discusions}
\subsection{Simulation and Analysis of the PPO Algorithm}\label{subsection_simulation_and_analysis_of_the_ppo_algorithm}
The initial states of both the EFV and the PFV for the training tasks are delineated in Table \ref{tab4}. By considering the combinations of all the proposed neural network parameters and learning rate, as displayed in Table \ref{tab1} and Table \ref{tab2}, 24 training missions with the corresponding configurations have been executed. Table \ref{tab5} presents the detailed training configurations, along with the corresponding evasion distance and residual velocity of the final episode, as achieved by the trained agent.

\begin{table}[tbp]
\centering
\caption{Initial states of both the EFV and the PFV}
\begin{tabular}{ccc}
\toprule
 \textbf{Parameter}                    &  \textbf{Meaning}                                                          &  \textbf{Value}                        \\
\midrule
 $d$                                   & \makecell{The relative distance \\ between the EFV and the PFV}            &  2000.0 m                              \\
 $\Delta t$                            & \makecell{The step time }                                                  &  0.01 s                                \\
 $\Delta {\underline\alpha}_{\textrm{cx}}$           & \makecell{The limited change of \\ ${\underline\alpha}_{\textrm{cx}}$ in a single training step}                                                  &  0.01$^{\circ}$                                  \\
 $\Delta {\gamma}_{\textrm{cx}}$           & \makecell{The limited change of \\ ${\gamma}_{\textrm{cx}}$ in a single training step}                                                  &  0.01$^{\circ}$                                  \\
 $v_{E}$                               & \makecell{The velocity of the EFV}                                         &  98.0640 m/s                            \\
 $a_{E}$                               & \makecell{The accelerate speed of the EFV}                                 &  0.9434 $m/s^2$                          \\
 ${\underline\alpha}_{\textrm{cx}}$    & \makecell{The composite angle of attack of the EFV}                        &  7.3443$^{\circ}$                        \\
 ${\gamma}_{\textrm{cx}}$              & \makecell{The angle of heel of the EFV}                                    &  -0.3755$^{\circ}$                       \\
 $v_{P}$                               & \makecell{The velocity of the PFV}                                         &  72.2426 m/s                           \\
 $a_{P}$                               & \makecell{The accelerate speed of the PFV}                                 &  0.5792 $m/s^2$                          \\
 ${\alpha}_{\textrm{cx}}$              & \makecell{The angle of attack of the PFV}                                  &  10.0898$^{\circ}$                       \\
 ${{\beta}_{\textrm{cx}}}$             & \makecell{The angle of sideslip of the PFV}                                &  0.2869$^{\circ}$                        \\
\bottomrule
\end{tabular}
\label{tab4}
\end{table}

\begin{table*}[tbp]
\caption{Training configurations and results of the PPO algorithm}
\centering
\begin{tabular}{cccccccc}
\toprule
\multirow{2}*{\textbf{Index}}      & \multicolumn{2}{c}{\textbf{Neural network}}      & \multicolumn{3}{c}{\textbf{Learning rate}}      & \multirow{2}*{\textbf{Evasion distance (m)}}      & \multirow{2}*{\textbf{Residual velocity (m/s)}}   \\
      & {\textbf{Architecture}}      & {\textbf{Activation function}}        & $k$         & ${l}_{B}$                  & \textbf{Steps}  \\
      \midrule
1     & [8, 64, 64, 2]          & ReLU                   & 2           & 0.001                      & 3500000                    & 0.8194              & 67.2922           \\
2     & [8, 128, 128, 2]        & ReLU                   & 2           & 0.001                      & 3500000                    & 2.9134              & 67.2072           \\
3     & [8, 256, 256, 2]        & ReLU                   & 2           & 0.001                      & 3500000                    & 5.0775              & 66.8269           \\
4     & [8, 64, 64, 64, 2]      & ReLU                   & 2           & 0.001                      & 3500000                    & 17.1040             & 67.5854           \\
5     & [8, 128, 128, 128, 2]   & ReLU                   & 2           & 0.001                      & 3500000                    & 52.1294             & 66.6821           \\
6     & [8, 256, 256, 256, 2]   & ReLU                   & 2           & 0.001                      & 3500000                    & \bf{30.8556 ($\star$)}   & \bf{67.2422 ($\star$)} \\
7     & [8, 64, 64, 2]          & ReLU                   & 2           & 0.0001                     & 3500000                    & 5.1875                 & 66.9888           \\
8     & [8, 128, 128, 2]        & ReLU                   & 2           & 0.0001                     & 3500000                    & 17.3354                 & 67.0435           \\
9     & [8, 256, 256, 2]        & ReLU                   & 2           & 0.0001                     & 3500000                    & 3.7665                 & 66.5762           \\
10    & [8, 64, 64, 64, 2]      & ReLU                   & 2           & 0.0001                     & 3500000                    & 19.2254                 & 67.0480           \\
11    & [8, 128, 128, 128, 2]   & ReLU                   & 2           & 0.0001                     & 3500000                    & 32.0025                 & 66.8794           \\
12    & [8, 256, 256, 256, 2]   & ReLU                   & 2           & 0.0001                     & 3500000                    & 50.2231                 & 66.5585           \\
13    & [8, 64, 64, 2]          & ReLU                   & 3           & 0.001                      & 3500000                    & 51.5623                 & 66.6671           \\
14    & [8, 128, 128, 2]        & ReLU                   & 3           & 0.001                      & 3500000                    & 13.2498                 & 66.8281           \\
15    & [8, 256, 256, 2]        & ReLU                   & 3           & 0.001                      & 3500000                    & 16.9923                 & 67.0745           \\
16    & [8, 64, 64, 64, 2]      & ReLU                   & 3           & 0.001                      & 3500000                    & 41.2203                 & 67.0789           \\
17    & [8, 128, 128, 128, 2]   & ReLU                   & 3           & 0.001                      & 3500000                    & 22.8876                 & 66.5946           \\
18    & [8, 256, 256, 256, 2]   & ReLU                   & 3           & 0.001                      & 3500000                    & 16.2531                 & 67.0824           \\
19    & [8, 64, 64, 2]          & ReLU                   & 3           & 0.0001                     & 3500000                    & 28.1975                 & 67.0743           \\
20    & [8, 128, 128, 2]        & ReLU                   & 3           & 0.0001                     & 3500000                    & 15.2231                 & 66.7912           \\
21    & [8, 256, 256, 2]        & ReLU                   & 3           & 0.0001                     & 3500000                    & 16.8831                 & 66.9802           \\
22    & [8, 64, 64, 64, 2]      & ReLU                   & 3           & 0.0001                     & 3500000                    & 31.1762                 & 66.5851           \\
23    & [8, 128, 128, 128, 2]   & ReLU                   & 3           & 0.0001                     & 3500000                    & 51.8064                 & 66.7531           \\
24    & [8, 256, 256, 256, 2]   & ReLU                   & 3           & 0.0001                     & 3500000                    & 17.2813                 & 67.0757           \\
\bottomrule
\end{tabular}
\label{tab5}
\end{table*}

According to the results in Table \ref{tab5}, the following observations are obtained.
\begin{enumerate}
\item
Under the given reward function, activation function and number of training steps, the performance of the trained agent is sensitive to the network architecture, the exponent $k$ and the starting value of the learning rate ${l}_{B}$. In other words, although theoretically the PPO algorithm is applicable to the multi-dimension action space problem, it is still difficult to find a sufficiently good solution to the problem \eqref{eq8} that has a large and uncertain $n$, if the hyper-parameters are not tuned properly.
\item
The optimal trained agent is specified by the parameter configuration of the 6th training mission, which achieves the final residual velocity of 67.2422 m/s and the final evasion distance of 30.8556 m. Although it is possible to obtain better results through a larger number of attempts, the effectiveness of this method is not guaranteed. Therefore, it is necessary to conceive another method to improve the achievable residual velocity.
\end{enumerate}

\begin{figure}[tbp]
\centering
  \subfloat[The evasion distance results during the training process.]{
    \label{fig9a}
    \includegraphics[width=0.95\linewidth]{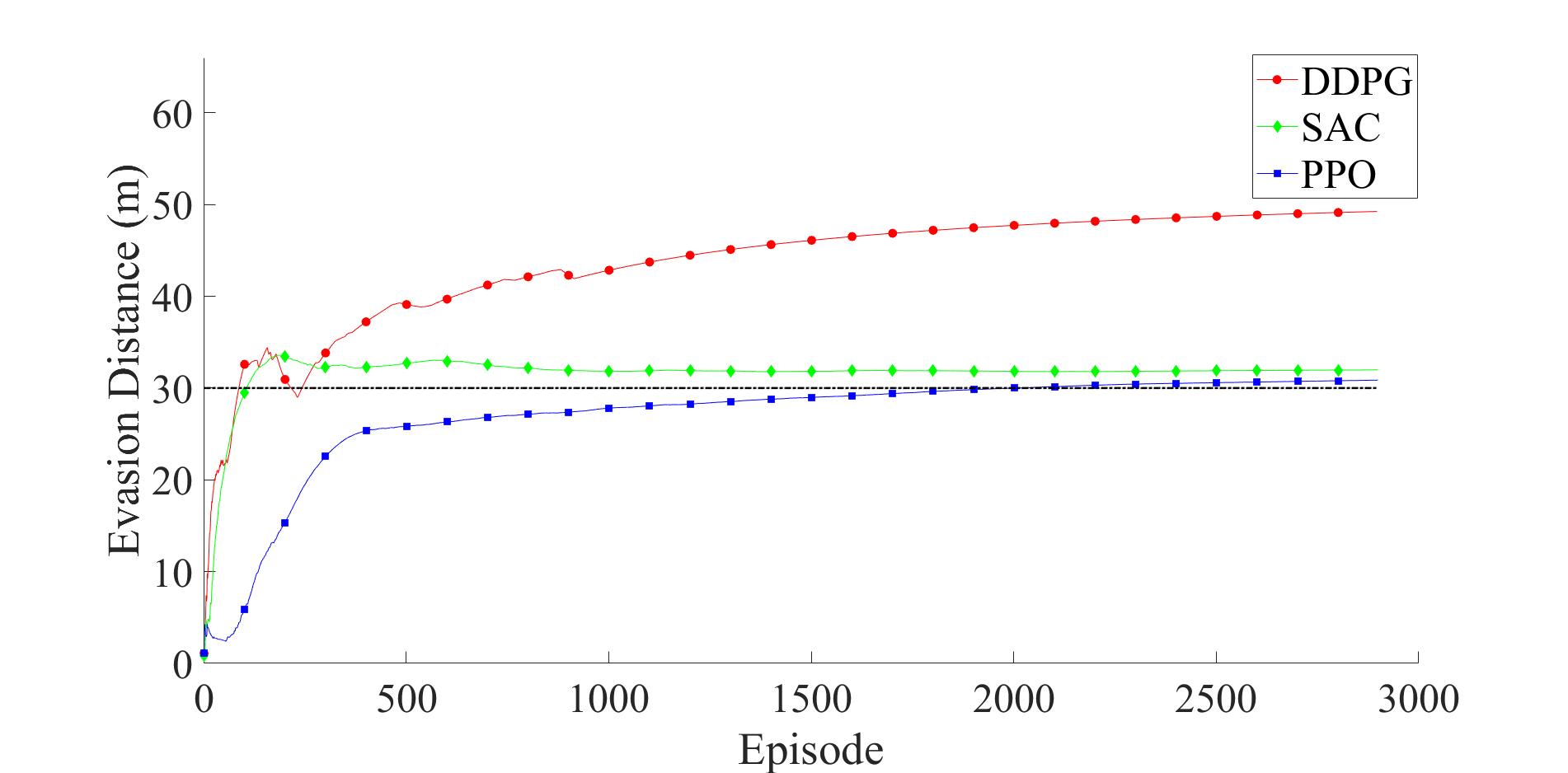}
  }
  \newline
  \subfloat[The residual velocity results during the training process.]{
    \label{fig9b}
    \includegraphics[width=0.95\linewidth]{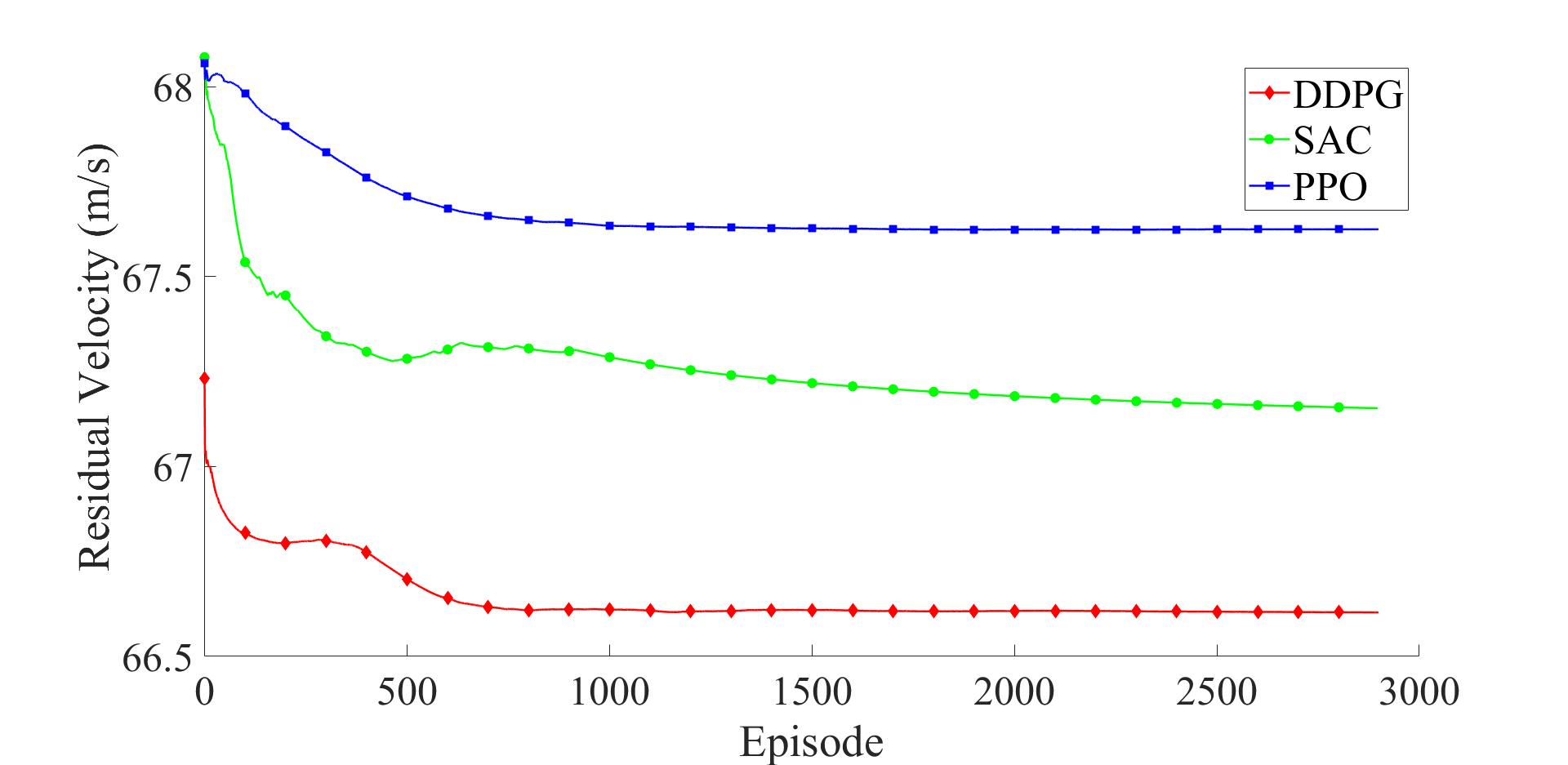}
  }
\caption{The best training results of the three algorithms.}
\label{fig9}
\end{figure}

To substantiate the PPO algorithm’s superiority, comparison of the top-performing outcomes yielded by the PPO algorithm, the deep deterministic policy gradient (DDPG) \cite{36} algorithm and the soft actor critic (SAC) \cite{37} algorithm is illustrated in Figure \ref{fig9}. As seen from Figure \ref{fig9a}, the evasion distances corresponding to the above three algorithms consistently exceed the safe threshold of 30.0m once they converge, indicating the EFV’s effective elusion of the PFV. Figure \ref{fig9b} reveals a general decrease in residual velocity as the number of episode increases, suggesting successful execution of training missions from the DRL perspective, while the PPO algorithm achieves the largest residual velocity among the three algorithms. It should be noted that the EFV’s energy is finite, and a portion of the energy is allocated to altering its flight trajectory, thereby enhancing its evasion distance. Consequently, the overall trend is that the residual velocity decreases with the number of episodes.

\subsection{Results of the ES-Enhanced PPO Algorithm}\label{subsection_result_of_the_es_enhanced_ppo_algorithm}
\begin{figure}[tbp]
 \centering
 \includegraphics[width=0.95\linewidth]{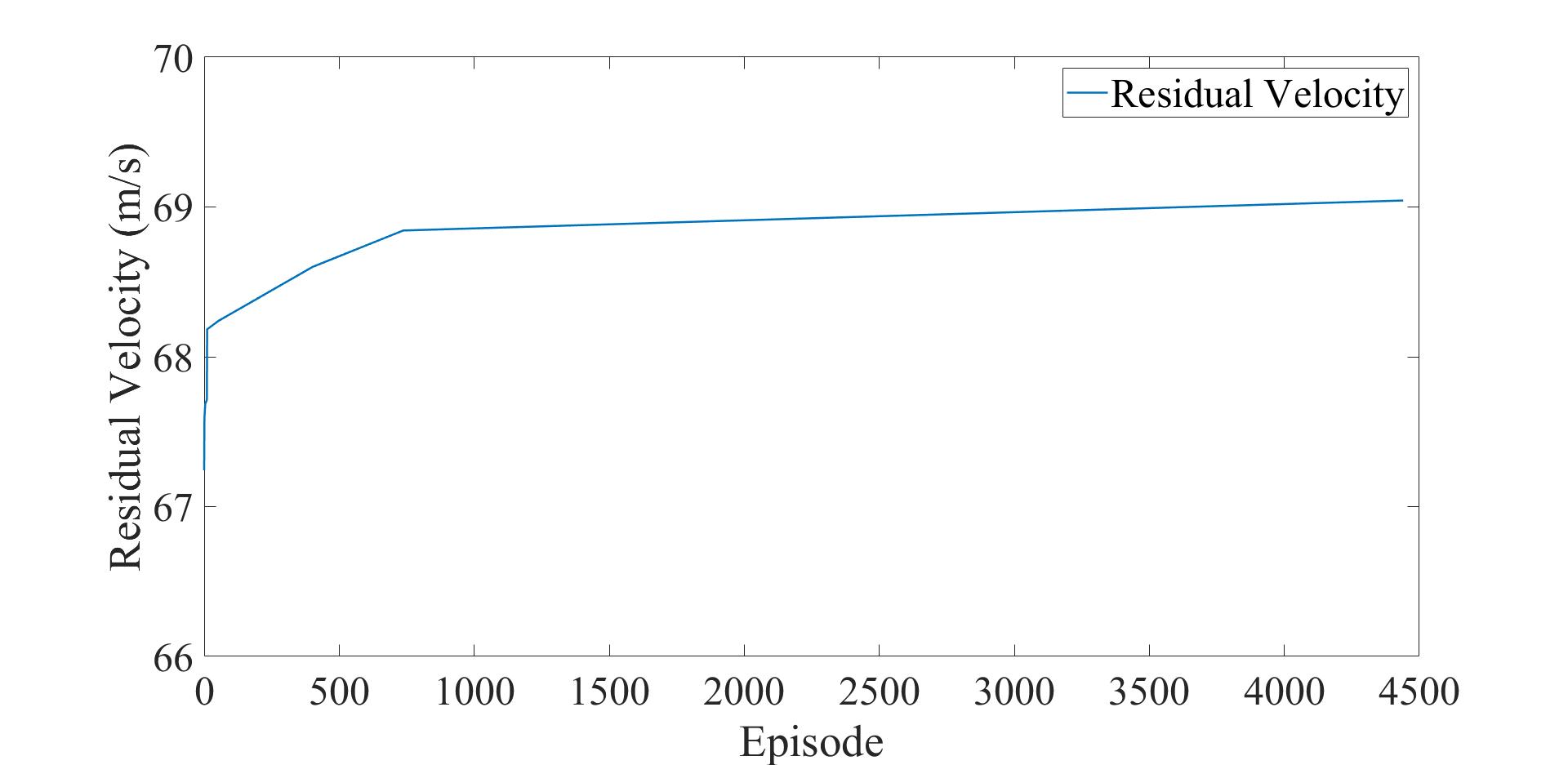}
 \caption{The residual velocity achieved by the successful evolution attempts in the ES-enhanced PPO algorithm.}
 \label{fig10}
\end{figure}

Since the 6th training task in Table \ref{tab5} achieves the best solution among the total 24 training tasks, it is set as the initial solution for the ES algorithm. By considering the parameter configurations of Table \ref{tab3}, 12 simulation setups and their individual final residual velocities and the evasion distances are presented in Table \ref{tab6}, where we can see that the best solution, i.e., the residual velocity of 69.0432 m/s with the evasion distance of 30.1036 m, occurs at the 8th simulation setup. Specifically, the record of successful evolution attempts in the 8th simulation setup is shown in Table \ref{tab7} and Figure \ref{fig10}.

\begin{table*}[tbp]
\caption{Simulation results of the ES-enhance PPO algorithm}
\centering
\begin{tabular}{ccccccc}
\toprule
\textbf{Index}   & \makecell{\textbf{Maximal allowed} \\ \textbf{number of generations}}    & \makecell{\textbf{Maximal allowed} \\ \textbf{number of seeds} \\ \textbf{in each generation}}   & \makecell{ \textbf{Variance of} \\ \textbf{normal distribution}} & \textbf{Evasion distance (m)}  & \textbf{Residual velocity (m/s)} & \makecell{\textbf{Better than the} \\ \textbf{initial solution?}}   \\
\midrule
1      &  50                 &  50           &  0.01             &  30.1324              &  67.5453                &  Yes                          \\
2      &  100                &  100          &  0.01             &  31.6741              &  67.2357                &  Yes                          \\
3      &  150                &  150          &  0.01             &  31.2246              &  68.7356                &  Yes                          \\
4      &  200                &  200          &  0.01             &  32.0994              &  68.1236                &  Yes                          \\
5      &  50                 &  50           &  0.1              &  30.8781              &  67.6724                &  Yes                          \\
6      &  100                &  100          &  0.1              &  31.9656              &  68.3261                &  Yes                          \\
7      &  150                &  150          &  0.1              &  32.3332              &  67.3714                &  Yes                          \\
8      &  200                &  200          &  0.1              &  \bf{30.1036 ($\star$)}    &  \bf{69.0432 ($\star$)}      &  Yes                          \\
9      &  50                 &  50           &  1.0              &  30.8556              &  67.2422                &  No                           \\
10     &  100                &  100          &  1.0              &  30.8556              &  67.2422                &  No                           \\
11     &  150                &  150          &  1.0              &  30.8556              &  67.2422                &  No                           \\
12     &  200                &  200          &  1.0              &  30.8556              &  67.2422                &  No                           \\
\bottomrule
\end{tabular}
\label{tab6}
\end{table*}

\begin{table}[tbp]
\centering
\caption{Evolution record under the 8th simulation setup}
\begin{tabular}{cccc}
\toprule
\textbf{Index}         & \makecell{\textbf{Evasion} \\ \textbf{distance (m)}}     &  \makecell{\textbf{Residual} \\ \textbf{velocity (m/s)}}   & \textbf{Note}                                                                 \\
\midrule
0                      &  30.8556                          &  67.2422              &  \makecell{The best trained \\ agent by only using \\ the PPO algorithm.}        \\
1                      &  30.5277                          &  67.5914              &                                                                               \\
4                      &  30.8377                          &  67.6842              &                                                                               \\
10                     &  30.4310                          &  67.7149              &                                                                               \\
11                     &  30.3886                          &  68.1833              &                                                                               \\
53                     &  30.5162                          &  68.2396              &                                                                               \\
401                    &  30.4767                          &  68.5995              &                                                                               \\
573                    &  30.4840                          &  68.7242              &                                                                               \\
737                    &  30.6595                          &  68.8428              &                                                                               \\
4440                   &  30.1036                          &  69.0432              &                                                                               \\
\bottomrule
\end{tabular}
\label{tab7}
\end{table}

\begin{figure}[tbp]
 \centering
 \includegraphics[width=0.95\linewidth]{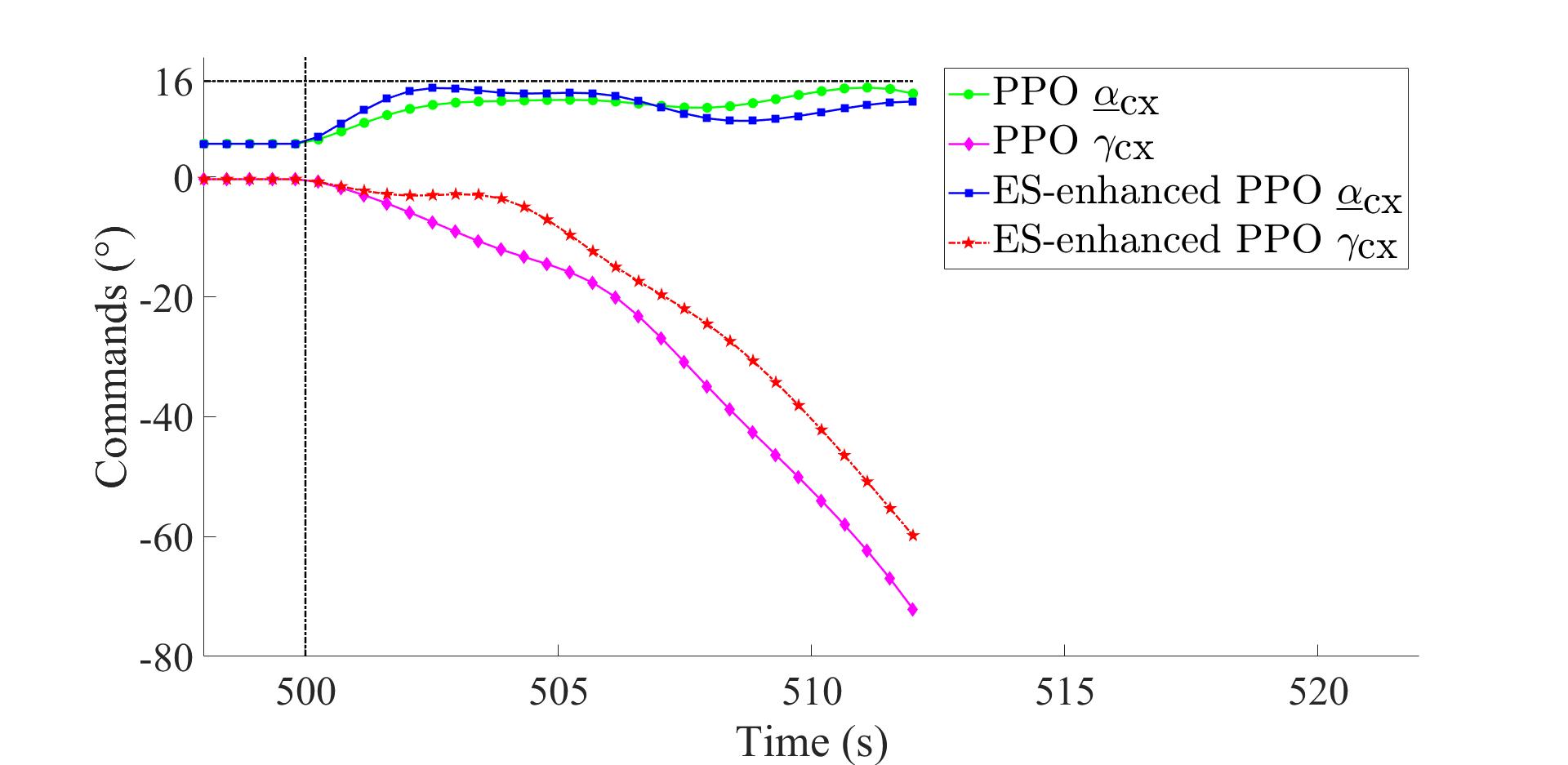}
 \caption{The guidance commands curves of the proposed algorithms.}
 \label{fig11}
\end{figure}

Based on the above simulation results, the following remarks are provided.
\begin{enumerate}
\item
The proposed ES-enhanced PPO algorithm can guarantee that the final result is no worse than the result obtained from using the PPO algorithm, because in the evolution process only the seed that achieves an residual velocity larger than the maximum residual velocity obtained so far is saved.
\item
The larger generation number and the larger seed number can make the ES-enhanced PPO algorithm explore the solution space more times, thus increasing the probability of achieving better solutions.
\item
The larger variance of normal distribution can make the ES-enhanced PPO algorithm explore a larger solution space on the basis of the initial solution, thus also increasing the probability of achieving better solutions. But under a given number of attempts, if the variance of normal distribution is too large, the probability of locating the best possible solution is reduced, and the ES-enhanced PPO algorithm may not be useful, as seen from the results of Table \ref{tab6}.
\item
It can be seen from Table \ref{tab7} that the residual velocity of the 4440th evolution attempt can reach 69.0432 m/s, which demonstrates the superior effectiveness of the ES-enhanced PPO algorithm, in comparison to the PPO algorithm. Furthermore, Figure \ref{fig11} presents the EFV's guidance commands, generated by both the PPO algorithm and the ES-enhanced PPO algorithm. It can be seen from Figure \ref{fig11} that ${\underline\alpha}_{\textrm{cx}}$ and ${\gamma}_{\textrm{cx}}$ changes significantly after 500.0 s, which represents the threshold of the appropriate operational timing of the algorithms.
\end{enumerate}

\section{Conclusions}\label{section_conclusions}
In this paper, we have considered the escape-and-pursuit problem of flight vehicles, where the EFV's residual velocity is expected to be maximized within the confines of an established evasion distance threshold. We assume that the EFV generates guidance commands with the aid of AI, while the PFV uses the conventional proportional navigation method. We reveal that in general it is difficult to find the analytical solution to the residual velocity maximization problem, subject to the evasion distance constraint. Since the guidance design problem considered constitutes a typical sequential decision problem and the results of decision in each step are from a multi-dimension continuous space, it is suitable to use the PPO algorithm. However, even if the reward function, the neural network architecture, the activation function and the learning rate are carefully designed, the PPO algorithm may still be trapped in the local optimum. To overcome this predicament, we propose an ES-enhanced PPO algorithm, where the ES operation is capable of guaranteeing an improved solution on the basis of the PPO algorithm's solution. Extensive simulation results demonstrate that the proposed guidance design method based on the PPO algorithm is capable of achieving a residual velocity of 67.24 m/s, higher than the residual velocities achieved by the benchmark soft actor-critic and deep deterministic policy gradient algorithms. Furthermore, the proposed ES-enhanced PPO algorithm outperforms the PPO algorithm by 2.7\%, achieving a residual velocity of 69.04 m/s.  The above results are indeed competitive. However, it is a more challenging and interesting scenario when both the EFV and the PFV take advantage of AI methods for guidance design, which will be studied in our future work.

%

\bibliographystyle{IEEEtran}

\begin{thebibliography}{[37]}

\bibitem{1} F. Chen, Y. L. Xiao, and W. C. Chen, ``Guidance based on zero effort miss for super-range exoatmospheric intercept,'' \emph{Acta Aeronautica et Astronautica Sinica}, vol. 30, no. 9, pp. 1583--1589, Sep. 2009.

\bibitem{2} Y. Q. Liu and N. M. Qi, ``A zero-effort miss distance-based guidance law for endoatmoshperic interceptor,'' \emph{Journal of Astronautics}, vol. 31, no. 7, pp. 1768--1774, Jul. 2010.

\bibitem{3} M. Cui and F. Geng, ``Solving singular two-point boundary value problem in reproducing kernel space,'' \emph{Journal of Computational and Applied Mathematics}, vol. 205, no. 1, pp. 6--15, Aug. 2007.

\bibitem{4} X. Y. Huang, D. Y. Wang, and Y. F. Guan, ``A linear quadratic optimal guidance method for lunar soft landing,'' \emph{Aerospace Control}, vol. 24, no. 6, pp. 11--16, Dec. 2006.

\bibitem{5} E. Garcia, D. W, Casbeer, and M. Pachter, ``Design and analysis of state-feedback optimal strategies for the differential game of active defense,'' \emph{IEEE Transactions on Automatic Control}, vol. 64, no. 2, pp. 553--568, Apr. 2018.

\bibitem{6} H. Liang, J. Wang, Y. Wang, L. Wang, and P. Liu, ``Optimal guidance against active defense ballistic missiles via differential game strategies,'' \emph{Chinese Journal of Aeronautics}, vol. 33, no. 3, pp. 978--989, Mar. 2020.

\bibitem{7} S. Liu, B. Yan, T. Zhang, X. Zhang, and J. Yan, ``Coverage-based cooperative guidance law for intercepting hypersonic vehicles with overload constraint,'' \emph{Aerospace Science and Technology}, vol. 126, article no. 107651, pp. 1--15, Jul. 2022.

\bibitem{8} A. Sinha, S. R. Kumar, and D. Mukherjee, ``Nonsingular impact time guidance and control using deviated pursuit,'' \emph{Aerospace Science and Technology}, vol. 115, article no. 106776, pp. 1--19, Aug. 2021.

\bibitem{9} W. Yu, W. Chen, Z. Jiang, W. Zhang, and P. Zhao, ``Analytical entry guidance for coordinated flight with multiple no-fly-zone constraints,'' \emph{Aerospace Science and Technology}, vol. 84, pp. 273--290, Jan. 2019.

\bibitem{10} A. Marchidan and E. Bakolas, ``Collision avoidance for an unmanned aerial vehicle in the presence of static and moving obstacles,'' \emph{Journal of Guidance, Control, and Dynamics}, vol. 43, no. 1, pp. 96--110, Jan. 2020.

\bibitem{11} Z. Xu, R. Wei, Q. Zhang, K. Zhou, and R. He, ``Obstacle avoidance algorithm for UAVs in unknown environment based on distributional perception and decision making,'' in \emph{Proc. IEEE Chinese Guidance, Navigation and Control Conference (CGNCC)}, Nanjing, China, Aug. 2016, pp. 1072--1075.

\bibitem{12} A. Mujumdar and R. Padhi, ``Reactive collision avoidance of using nonlinear geometric and differential geometric guidance,'' \emph{Journal of Guidance, Control, and Dynamics}, vol. 34, no. 1, pp. 303--310, Jan. 2011.

\bibitem{13} I. Goodfellow, Y. Bengio, and A. Courville, \emph{Deep Learning.} Cambridge, MA, USA: MIT Press, 2016.

\bibitem{14} R. S. Sutton and A. G. Barto, \emph{Reinforcement Learning.} Beijing, China: Publishing House of  Electronics Industry, 2019.

\bibitem{15} J. Schulman, S. Levine, P. Moritz, M. I. Jordan, and P. Abbeel, ``Trust region policy optimization,'' 2015. [Online]. Available: https://arxiv.org/abs/1502.05477v3.

\bibitem{16} Y. Wang, H. He, C. Wen and X. Tan, ``Truly proximal policy optimization,'' 2019. [Online]. Available: https://arxiv.org/abs/1903.07940v1.

\bibitem{17} H. G. Beyer and H. P. Schwefel, ``Evolution strategies--a comprehensive introduction,'' \emph{Natural Computing}, vol. 1, pp. 3--52, Mar. 2002.

\bibitem{18} D. Silver, A. Huang, C. J. Maddison \emph{et al.}, ``Mastering the game of Go with deep neural networks and tree search,'' \emph{Nature}, vol. 529, no. 7587, pp. 484--489, Jan. 2016.

\bibitem{19} D. Silver, J. Schrittwieser, K. Simonyan \emph{et al.}, ``Mastering the game of Go without human knowledge,'' \emph{Nature}, vol. 550, no. 7676, pp. 354--359, Oct. 2017.

\bibitem{20} O. Vinyals, I. Babuschkin, W. M. Czarnecki \emph{et al.}, ``Grandmaster level in StarCraft II using multi-agent reinforcement learning,'' \emph{Nature}, vol. 575, no. 7782, pp. 350--354, Oct. 2019.

\bibitem{21} D. Ye, G. Chen, P. Zhao \emph{et al.}, ``Supervised learning achieves human-level performance in MOBA games: A case study of honor of kings,'' \emph{IEEE Transactions on Neural Networks and Learning Systems}, vol. 33, no. 3, pp. 908--918, Mar. 2022.

\bibitem{22} Q. H. Zhang, B. Q. Ao, and Q. X. Zhang, ``Q-learning reinforcement learning guidance law,'' \emph{Systems Engineering and Electronics}, vol. 42, no. 2, pp. 414--419, Feb. 2020.

\bibitem{23} C. Liang, W. Wang, Z. Liu, C. Lai, and B. Zhou, ``Learning to guide: Guidance law based on deep meta-learning and model predictive path integral control,'' \emph{IEEE Access}, vol. 7, pp. 47\,353--47\,365, Apr. 2019.

\bibitem{24} A. Rajagopalan, ``Intelligent missile guidance using artificial neural networks,'' \emph{Journal of Artificial Intelligence Research}, vol. 4, no. 1, pp. 60--76, Mar. 2015.

\bibitem{25} B. Gaudet, R. Furfaro, and R. Linares, ``Reinforcement learning for angle-only intercept guidance of maneuvering targets,'' \emph{Aerospace Science and Technology}, vol. 99, article no. 105746, pp. 1--10, Apr. 2020.

\bibitem{26} E. Meyer, A. Heiberg, A. Rasheed, and O. San, ``COLREG-compliant collision avoidance for unmanned surface vehicle using deep reinforcement learning,'' \emph{IEEE Access}, vol. 8, pp. 165\,344--165\,364, Sep. 2020.

\bibitem{27} K. Wu and P. Yao, ``Obstacle avoidance for AUV by Q-learning based guidance vector field,'' in \emph{Proc. 3rd International Conference on Unmanned Systems (ICUS)}, Harbin, China, Nov. 2020, pp. 702--707.

\bibitem{28} Z. Li, X. Sun, C. Hu, G. Liu, and B. He, ``Neural network based online predictive guidance for high lifting vehicles,'' \emph{Aerospace Science and Technology}, vol. 82--83, pp. 149--160, Nov. 2018.

\bibitem{29} L. Cheng, F. Jiang, Z. Wang, and J. Li, ``Multiconstrained real-time entry guidance using deep neural networks,'' \emph{IEEE Transactions on Aerospace and Electronic Systems}, vol. 57, no. 1, pp. 325--340, Feb. 2020.

\bibitem{30} C. Peng, H. Zhang, Y. He, and J. Ma, ``State-following-kernel-based online reinforcement learning guidance law against maneuvering target,'' \emph{IEEE Transactions on Aerospace and Electronic Systems}, vol. 58, no. 6, pp. 5784--5797, Dec. 2022.

\bibitem{31} C. Wang, D. Deng, L. Xu, and W. Wang, ``Resource scheduling based on deep reinforcement learning in UAV assisted emergency communication networks,'' \emph{IEEE Transactions on Communications}, vol. 70, no. 6, pp. 3834--3848, Jun. 2022.

\bibitem{32} R. Ding, F. Gao, and X. S. Shen, ``3D UAV trajectory design and frequency band allocation for energy-efficient and fair communication: A deep reinforcement learning approach,'' \emph{IEEE Transactions on Wireless Communications}, vol. 19, no. 12, pp. 7796--7809, Dec. 2020.

\bibitem{33} X. F. Qian, \emph{Missile Flight Aerodynamics.} Beijing, China: Beijing Institute of Technology Press, 2014.

\bibitem{34} J. Schulman, F. Wolski, P. Dhariwal, A. Radford, and O. Klimov, ``Proximal policy optimization algorithms,'' 2017. [Online]. Available: http://arxiv.org/abs/1707.06347.

\bibitem{35} B. Fu, ``Guidance design for the interception of hypersonic boost glide vehicles,'' Ph.D. dissertation, College of Astronautics, Northwestern Polytechnical University, 2019.

\bibitem{36} T. P. Lillicrap, J. J. Hunt, A. Pritzel \emph{et al.}, ``Continuous control with deep reinforcement learning,'' 2019. [Online]. Available: https://arxiv.org/abs/1509.02971

\bibitem{37} T. Haarnoja, A. Zhou, K. Hartikainen, \emph{et al.}, ``Soft actor-critic algorithms and applications,'' 2019.  [Online]. Available: https://arxiv.org/abs/1812.05905

\end{thebibliography}

\begin{IEEEbiography}[{\includegraphics[width=1in,height=1.25in,clip,keepaspectratio]{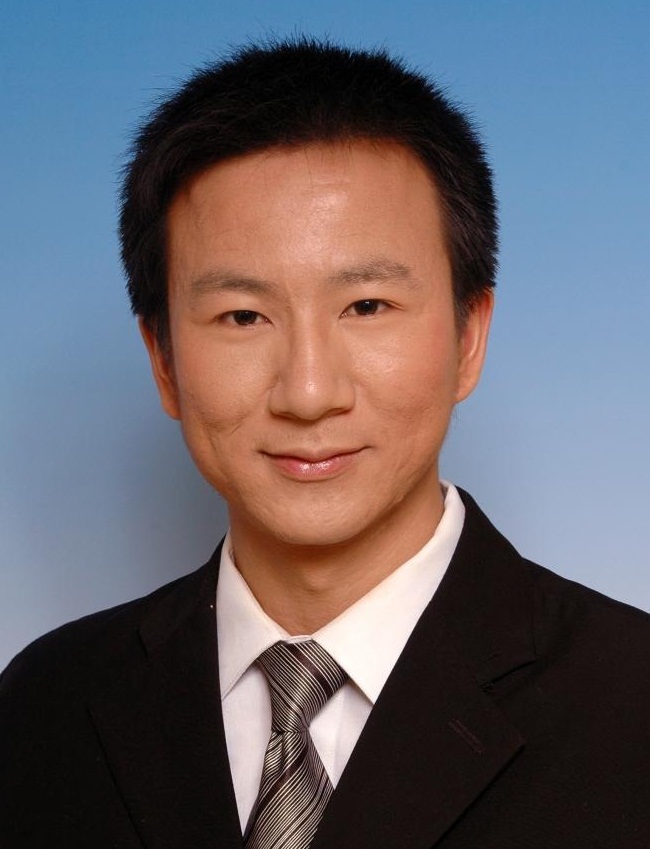}}]{Shaoshi Yang} received the B.Eng degree in information engineering from Beijing University of Posts and Telecommunications (BUPT), Beijing, China, in 2006, and the Ph.D. degree in electronics and electrical engineering from University of Southampton, U.K., in 2013. From 2008 to 2009, he was a researcher of WiMAX standardization with Intel Labs China. From 2013 to 2016, he was a Research Fellow with the School of Electronics and Computer Science, University of Southampton. From 2016 to 2018, he was a Principal Engineer with Huawei Technologies Co. Ltd., where he made significant contributions to the company's products and solutions on 5G base stations, wideband IoT, and cloud gaming/VR. He is currently a Full Professor with BUPT. His research expertise includes 5G wireless networks, massive MIMO, iterative detection and decoding, mobile ad hoc networks, distributed artificial intelligence (AI), and cloud gaming/VR. He was a Guest Member of the Isaac Newton Institute for Mathematical Sciences, Cambridge University, Cambridge, U.K. He received the Deans Award for Early Career Research Excellence from University of Southampton in 2015, the Huawei President Award of Wireless Innovations in 2018, the IEEE Technical Committee on Green Communications and Computing (TCGCC) Best Journal Paper Award in 2019, and the IEEE Communications Society Best Survey Paper Award in 2020. He is an editor for IEEE Systems Journal, IEEE Wireless Communications Letters, and Signal Processing (Elsevier). He was also an invited international reviewer of the Austrian Science Fund (FWF).
\end{IEEEbiography}

\end{document}